\documentclass[fontset=none,10pt,a4paper]{ctexart}

\usepackage{style}
\aihsetupmain
\ctexset{abstractname = Abstract, bibname = References}
\title{AI Historian: Helping historians organize and verify person-centred temporal clues from dispersed historical narratives}
\author{
Yifeng Lu\textsuperscript{1,*} \quad
Zijie Yang\textsuperscript{1,*} \quad
Jie Li\textsuperscript{2,*} \quad
Qingkai Min\textsuperscript{1} \quad
Yue Zhang\textsuperscript{1,\textdagger}\\[0.5em]
\textsuperscript{1}Westlake University\\
\textsuperscript{2}Peking University\\[0.25em]
\textsuperscript{*}These authors contributed equally to this work.\\
\textsuperscript{\textdagger}Corresponding author.\\
Emails: \{luyifeng, yangzijie, minqingkai, zhangyue\}@westlake.edu.cn; licuanlin@pku.org.cn
}

\date{}

\begin{document}

\maketitle
\aihpagelogo

\begin{abstract}
History is not preserved in complete, continuous form. Accounts of a person's activities, relationships and historical contexts are scattered across texts, chapters and narrative perspectives; historians must retrieve, identify and compare these materials to reconstruct temporal sequences and verify them against sources. Here we present AI Historian (AIH), an AI agent system that helps historians organize person--time evidence from dispersed biographical narratives. It takes source sentences as evidence units, identifies people and temporal cues, verifies candidate cross-text associations and infers comparable temporal ranges while preserving traceable source-text evidence. We evaluated AIH on six \textit{Shiji} cases concerning Liu Bang, Xiang Yu and Xiao He. AIH Agent achieved a temporal-localization MicroIoU of 86.2\%, compared with 81.3\% for human-only annotation and 17.1\% for direct large-language-model prompting; it required about 14\,min, versus 1\,h 32\,min for human-only annotation. We further applied AIH to the Twenty-Four Histories and other ancient Chinese histories, ancient Japanese and Korean histories, and modern and contemporary historical materials, and released the results through Westlake Historian. These results indicate that AIH can reduce the cost of organizing historical materials at scale while turning connections obscured by chapter-based narration into traceable, revisable research questions for collaborative testing.
\end{abstract}

\section{Introduction}

Historical research often begins by reconnecting dispersed materials. Information about the time, participants, actions and consequences of the same person or event may be preserved in different documents, chapters and narrative perspectives: one source may provide a date, another an action or relationship, and different materials may corroborate or diverge from one another. Only through sustained retrieval, identification and comparison can historians reconnect these fragments and identify historical leads that warrant further inquiry\cite{stone1971prosopography,pasin2015factoid}. This is not simply a problem of information retrieval. Historical narratives confer meaning on dispersed facts through selection, arrangement and interpretation\cite{white1973metahistory,ankersmit1983narrativelogic}; source criticism requires tracing textual provenance, distinguishing relations among materials and preserving the process of evaluation\cite{chen1980shiyuan,liang1922historymethod}; and human actions must be understood across different temporal scales, including events, processes and the longue durée\cite{braudel1958longueduree}. Yet conventional textual notes and manual compilation make it difficult to preserve systematically the complete evidentiary and inferential path from material selection and temporal judgement to historical interpretation, especially when evidence dispersed across multiple biographical narratives must be verified item by item.

This challenge is particularly acute in histories written in the annals-biographies format. The annalistic format arranges accounts by year and month, whereas the annals-biographies format distributes related material across chapters devoted to different people, scattering evidence about the same person or event\cite{kurz2015traditional,twitchett1992official}. The \textit{Shiji} (commonly translated as \textit{Records of the Grand Historian}) is China's first comprehensive history in the annals-biographies format. Its account reaches back to the legendary past and extends to around 94 BCE; it became an important model for later dynastic histories and a foundational source for the study of early China\cite{hardy1999worlds,li2020shiji}. Our cases focus on the transition from the fall of the Qin (China's first unified imperial dynasty) to the early Han (when the Han consolidated the unified imperial order and profoundly shaped subsequent Chinese history): Liu Bang founded the Han empire that succeeded the Qin, Xiang Yu was his principal rival for supremacy during the Chu--Han Contention, and Xiao He was a key administrator and organizer of the rear in Liu Bang's camp. Accounts of Liu Bang appear not only in the \textit{Basic Annals of Gaozu}, but also in the \textit{Basic Annals of Xiang Yu}, the \textit{Hereditary House of Chancellor Xiao} and other chapters (\supfigref{supp-fig:supp-biographical-fragmentation}). Only by retrieving, temporally juxtaposing and comparing these dispersed passages can researchers reconstruct the sequence of Liu Bang's activities, examine relations among the actions of different people and identify corroboration, divergence and omission across accounts.

The form adopted in the \textit{Shiji} continued in later dynastic histories. The Twenty-Four Histories, beginning with the \textit{Shiji}, constitute the core corpus of China's traditional official historiography and collectively span from the legendary age of the Yellow Emperor to the end of the Ming dynasty\cite{kurz2015traditional}. Major Japanese and Korean histories, including the \textit{Dai Nihonshi} and the \textit{Goryeosa}, likewise use biographies as an important means of organizing historical materials\cite{webb1960dai,kim2014goryeosa}. Other historiographical traditions also organize historical narratives around prominent lives; Plutarch's \textit{Parallel Lives}, for example, presents Greek and Roman figures in biographical form\cite{duff2002plutarch}. Across these traditions, researchers face a common problem: how to identify people in dispersed records, reconstruct their activities and relationships, and preserve the original evidence supporting each judgement.

Traditional biographical chronologies provide an established reference for organizing such dispersed evidence. Chinese biographical chronologies typically arrange a historical figure's actions, associations, writings and related events as a continuous sequence of entries, whereas extended biographical chronologies can accommodate richer source excerpts and editorial judgements (\supfigref{supp-fig:supp-nianpu-forms})\cite{shi2020nianpu,jia2020nianpu,sang2019changbian}. Other historiographical traditions similarly reconstruct individual lives through dated documents and correspondence, as in \textit{The Life and Letters of Thomas Henry Huxley}\cite{huxley1900life}. High-quality compilations generally identify their sources and explain the selection of materials and temporal judgements\cite{jia2020nianpu,wang2022liangnianpu}. Nevertheless, material selection, temporal judgement and the compiled result remain primarily dispersed within prose, with few explicit cross-references or correspondences among them, making item-level tracing, calculation and comparison difficult. We therefore take traditional biographical chronology as a reference and examine whether AIH can reproduce historians' practice of chronology compilation across a larger corpus of histories in the annals-biographies format: identifying episodes in individuals' lives from dispersed records, arranging and comparing the relevant texts chronologically, and making every judgement traceable to its source-text evidence. Biographical chronology thus provides a concrete point of entry into the broader problem of organizing historical literature.

Digital history has developed historical knowledge graphs, biographical databases and social networks that provide structured representations of people, places, events and relationships\cite{chen2022cbdb,hyvonen2019biographysampo,hyvonen2016warsampo}. Previous studies have addressed historical date recognition, event ordering and relative timelines, temporal reasoning by language models and event knowledge graphs\cite{strotgen2014extending,leeuwenberg2018temporal,chambers2014dense,ning2018multiaxis,wang2024tram,guan2023eventkg}. Although these tasks provide components needed to transform fragmented biographical narratives into person-centred timelines, they do not integrate dispersed judgements into an item-level traceable result, and performing this integration entirely by hand is too time-consuming to be practical at corpus scale.

Large language models and agent technologies have been applied to scientific hypothesis generation, chemical reasoning and experimental planning, and to searches for new mathematical constructions and algorithms using programmatic evaluation, demonstrating their potential to organize complex research tasks into sequential steps\cite{wang2024scimon,yang2024opendomain,yang2025moosechem,bran2024chemcrow,boiko2023autonomous,romeraparedes2024funsearch,novikov2025alphaevolve,gottweis2026coscientist}. These advances create new possibilities for material collection, temporal judgement and cross-text verification in historical research. Yet the position of evidence in long contexts affects model performance\cite{liu2024lost}; model temporal reasoning remains below human performance\cite{wang2024tram}; and citation support and self-evaluation of generated content are not consistently reliable\cite{gao2023citations,si2025novelideas}. Historical research therefore still requires material retrieval, temporal judgement and evidence verification to be separated into inspectable steps, with sources retained for review by historians.

To address the dispersal of records about historical figures across texts and the difficulty of organizing and verifying their temporal relations, we developed AI Historian (AIH) by combining large language models with multi-agent technology. AIH's principal contribution is a common representation of temporal evidence that organizes dispersed person-centred narratives from the Twenty-Four Histories and other works of ancient Chinese history, as well as Japanese and Korean histories, on a unified timeline, transforming them into comparable, chronologically organized person--time evidence. As an AI agent system designed to achieve this goal (\figref[a]{fig1}), AIH takes source sentences as evidence units and preserves source provenance, inferential paths and uncertainty. A sentence can be annotated with multiple people and enter each person's evidence collection (\figref[b]{fig1}). The system then produces candidate temporal ranges from each person's evidence collection (\figref[c]{fig1}) and aligns ranges for different people on a shared temporal axis (\figref[d]{fig1}). Historians can thereby compare temporal overlaps among people's activities and return to the original texts to assess their historical significance.

To evaluate the accuracy and efficiency with which AIH organizes person--time evidence, we compared it with human-only annotation and direct large-language-model prompting across six \textit{Shiji} cases. AIH Agent achieved a temporal-localization accuracy of 86.2\%, exceeding 81.3\% for human-only annotation and 17.1\% for direct large-language-model prompting. It completed the same temporal annotation task in approximately 14\,min, compared with 1\,h 32\,min for human-only annotation.

To enable historians to retrieve, verify and revise AIH-organized temporal evidence continuously, we further implemented it as the publicly accessible Westlake Historian platform (\url{https://westlakehistorian.com}). The platform currently presents examples covering ancient Chinese, Japanese and Korean histories as well as modern and contemporary historical sources, and supports comparison of temporal evidence across texts and people. AIH and the platform thus connect the organization of dispersed sources, the identification of historical leads and verification against original texts into a traceable, revisable and shareable research process, providing a practical approach to AI-assisted, cross-text research on historical figures.

\clearpage

\begin{figure}[H]
  \centering
  \includegraphics[width=\textwidth]{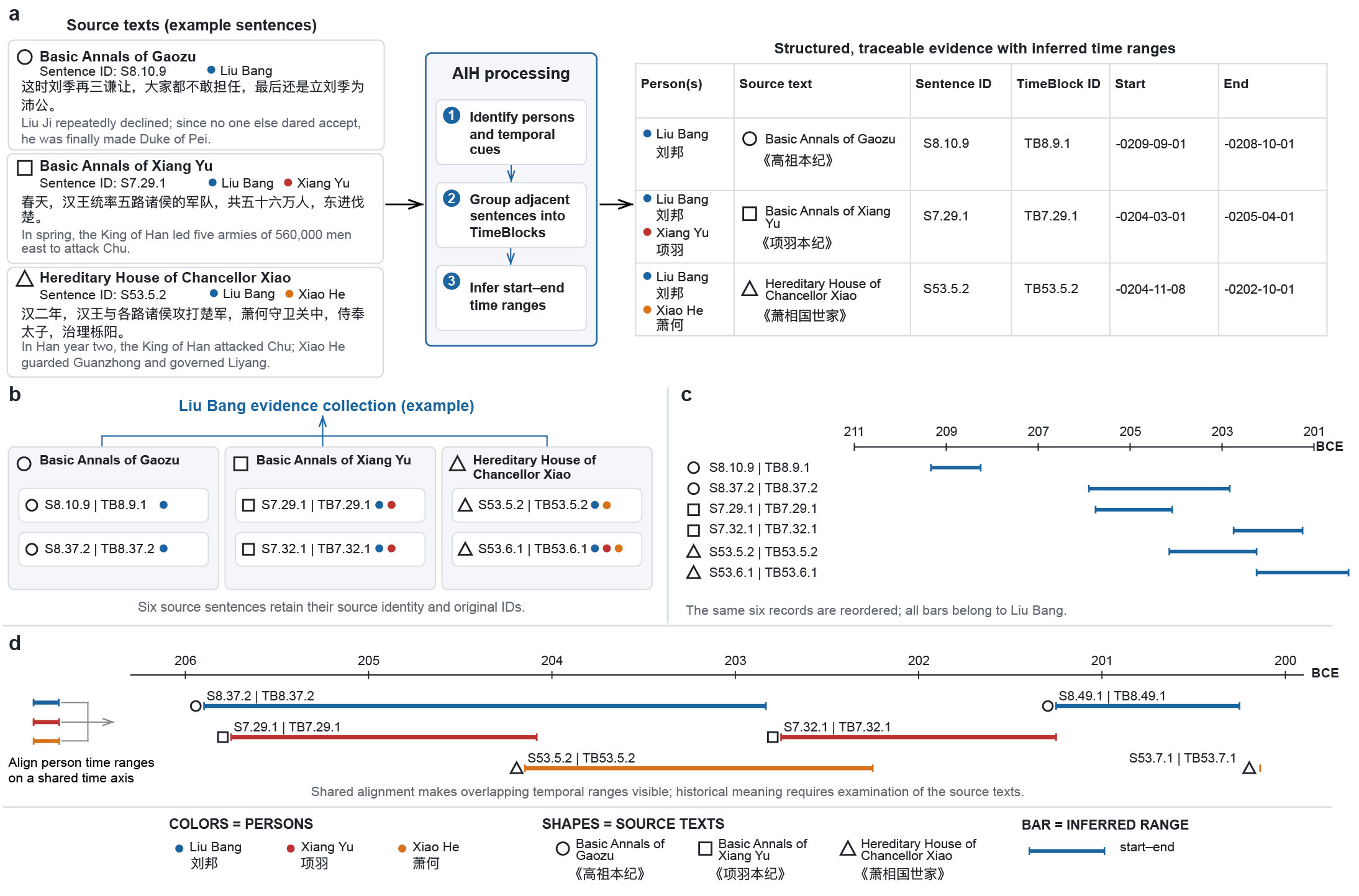}
  \caption[AIH organizes dispersed source sentences into person-centred timelines]{AIH organizes dispersed source sentences into person-centred timelines and aligns them on a shared temporal axis. \textbf{a,} From example sentences in the \textit{Basic Annals of Gaozu}, the \textit{Basic Annals of Xiang Yu} and the \textit{Hereditary House of Chancellor Xiao}, AIH identifies people and temporal cues, groups adjacent sentences sharing the same temporal context into TimeBlocks, and records inferred ranges using start and end dates; the bilingual table on the right retains the person, source chapter, sentence number and TimeBlock number. \textbf{b,} Six relevant sentences concerning Liu Bang are assembled from the three source texts while retaining their source identities and original numbering. Coloured dots on the right of each sentence card mark associated people, so a single sentence can be linked to Liu Bang, Xiang Yu and/or Xiao He. \textbf{c,} The six pieces of evidence are reordered according to their inferred temporal ranges to produce a single-person timeline for Liu Bang; shapes continue to denote their respective sources. \textbf{d,} Candidate ranges generated separately for Liu Bang, Xiang Yu and Xiao He are aligned on a shared temporal axis, making temporal overlaps directly visible; the diagram on the left shows their convergence from person-specific ranges into a common temporal coordinate system. Researchers assess the historical significance of these overlaps against the source texts. Colours denote people, shapes denote source texts and bars denote inferred temporal ranges.}
  \label{fig1}
\end{figure}

\section{Results}

\subsection{Bringing people scattered across historical texts back onto the same timeline}

In histories written in the annals-biographies format, the same historical process is often divided among chapters devoted to different people. Historical figures do not enter the record along a single timeline. The same war may appear as advances and retreats in one biography, as the movement of provisions in another and only as an imprecise date in a third. Historians must bring these misaligned narratives back together before they can see how different people acted within the same historical process. This is the problem AIH addresses: not writing a seamless story for the historian, but organizing dispersed materials into temporal evidence from which every claim can be traced back to the source sentence.

We used AIH to process 25 collections in Westlake Historian. These comprise 75 texts selected from 26 historical works and one collection of materials on the modern history of artificial intelligence, encompass 73 people and contain approximately 1.202 million characters. With tasks executed in parallel, AIH processed these materials in approximately 2\,d; we estimate that one researcher would require about 169 eight-hour working days to organize material at the same scale. At the public API price for Qwen3.6-35B-A3B, model calls would cost approximately RMB 1,000--2,000 (US\$147--295). Thus, dispersed material that would otherwise require months of manual organization can be transformed in about two days into candidate evidence for further verification by historians (\suptabref{supp-tab:supp-platform-coverage}).

Scale, however, is not the result in itself. Figure~\ref{fig1} uses Liu Bang to show how the historical materials behind these numbers are brought back into contact. His activities are dispersed across the \textit{Basic Annals of Gaozu}, the \textit{Basic Annals of Xiang Yu} and the \textit{Hereditary House of Chancellor Xiao}, and are embedded variously in narratives centred on a wartime rival, Liu Bang's own career and administration in the rear. AIH retrieves the relevant sentences from all three biographies and places them on a shared temporal axis, allowing researchers to view the activities of Liu Bang, Xiang Yu and Xiao He during the same period and to trace any temporal range back to the person, chapter, source sentence and basis of inference. The result is not a narrative summary detached from the texts, but an evidence chain that can be questioned and revised.

\begin{figure}[H]
  \centering
  \includegraphics[width=1\textwidth]{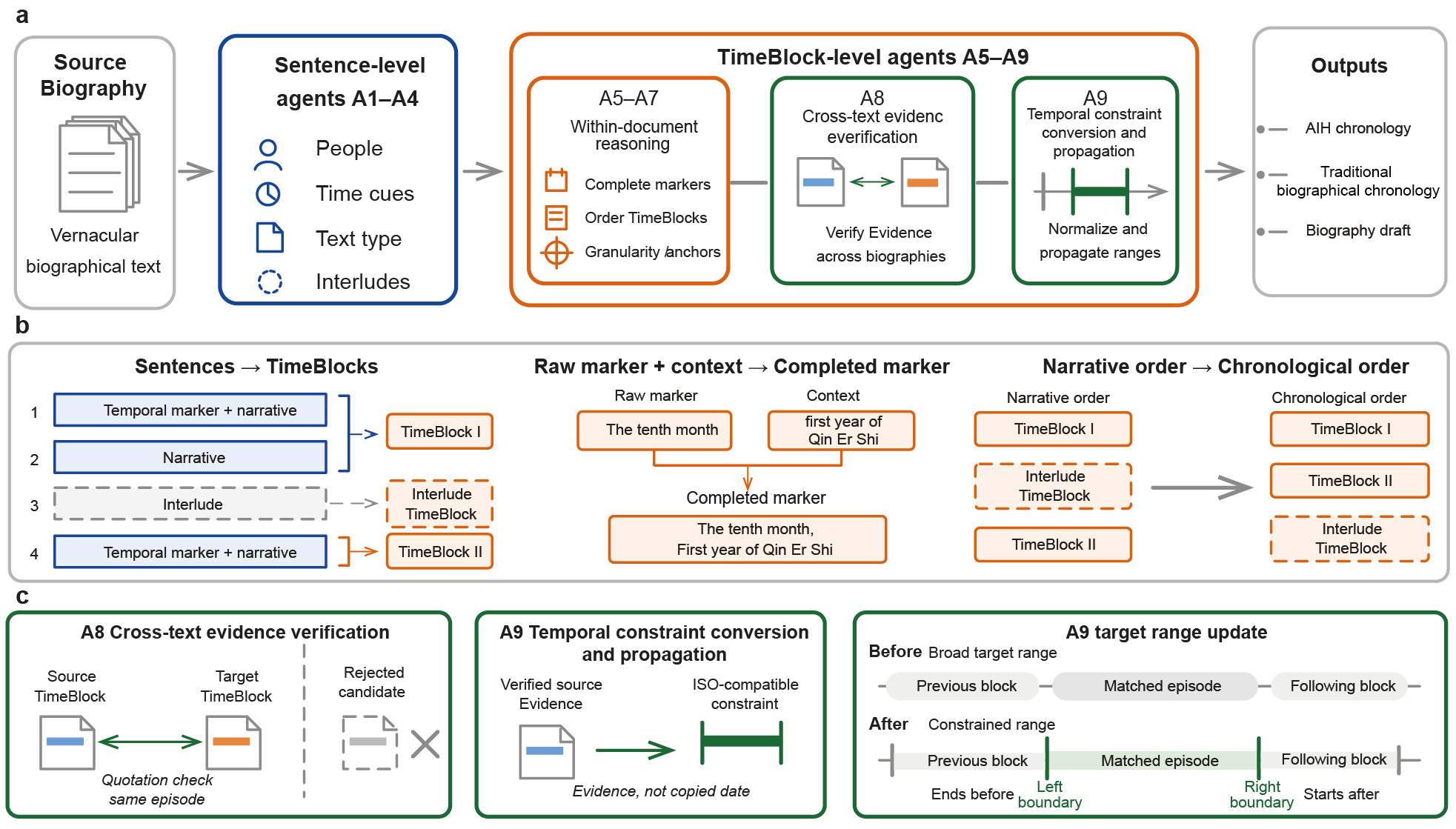}
  \caption[AIH Agent architecture and TimeBlock reasoning]{AIH Agent architecture and TimeBlock-level temporal reasoning. \textbf{a,} AIH Agent takes biographies from histories in the annals-biographies format as input. Sentence-level agents (A1--A4) identify people, extract temporal expressions from the original text, determine narrative functions and mark temporal shifts caused by retrospective passages. TimeBlock-level agents (A5--A9) then perform within-document temporal reasoning, cross-biography evidence verification and normalized temporal-coordinate generation, producing traceable person--time evidence; the biographical chronology and temporal-range views are system outputs. \textbf{b,} A TimeBlock groups sentences governed by the same temporal marker or temporal context into a sortable unit, supporting completion of incomplete temporal markers, determination of temporal granularity and anchors, and reordering from narrative into chronological order. \textbf{c,} Before temporal information is transferred, A8 assesses whether the selected source and target TimeBlocks provide sufficient evidence to support the same event or historical phase. Relations that meet the propagation criteria enter A9; other candidates are retained for human review. A9 converts verified temporal evidence from the source TimeBlock into normalized boundary constraints that narrow the temporal range of the target TimeBlock. These internal coordinates provide the system's standardized representation for ordering and scoring; their serialization is defined in \suppsecref{supp-sec:supp-nine-agent-workflow}. Detailed architecture and methods are provided in \supfigrangeref{supp-fig:supp-aih-agent-architecture}{supp-fig:supp-cross-textual-propagation}.}
  \label{fig:aih-multi-agent-pipeline}
\end{figure}

The key to this transformation is not to ask a large language model to extract, consolidate and date the material in a single step, but to leave an inspectable basis for every step. AIH preserves source sentences, groups narratives that share a temporal context, identifies shifts in order caused by retrospective passages and verifies that two biographies are indeed related before allowing temporal information in one to constrain the other. Only verified evidence is converted into standardized temporal ranges (\figref{fig:aih-multi-agent-pipeline}). The division of labour among agents is detailed in \suppsecref{supp-sec:supp-nine-agent-workflow}, the TimeBlock variables and ordering rules in \suppsecref{supp-sec:supp-data-representation}, and the complete architecture and cross-text propagation in \supfigrangeref{supp-fig:supp-aih-agent-architecture}{supp-fig:supp-cross-textual-propagation}.

\subsection{Accurate and efficient reconstruction of person--time evidence is a multifaceted challenge}

Placing dispersed materials on a common timeline is only the first step; the more important question is whether historical figures can be placed at the correct time. We tested this using six cases concerning Liu Bang, Xiang Yu and Xiao He in the \textit{Shiji}. Across 244 sentence-level temporal ranges, AIH Agent achieved a MicroIoU of 86.2\%, exceeding human-only annotation (81.3\%) and direct large-language-model prompting (17.1\%); its cumulative elapsed time was 13\,min 56\,s, compared with 1\,h 32\,min for human-only annotation (\figref[b,c]{fig:evaluation-results}). Direct prompting required only 2\,min 28\,s, but frequently placed sentences in the wrong historical period. This contrast shows that the difficulty is not in producing a date quickly, but in determining which parts of a narrative are governed by a temporal cue, restoring an order disrupted by retrospective passages and verifying whether different biographies indeed describe the same event. To locate the source of these errors, we asked three further questions: was the temporal cue in the sentence interpreted correctly, was the sequence within a biography restored, and could materials across biographies be aligned to the same historical period on the basis of evidence (\figref[a]{fig:evaluation-results})?

\begin{figure}[H]
  \centering
  \includegraphics[width=\textwidth]{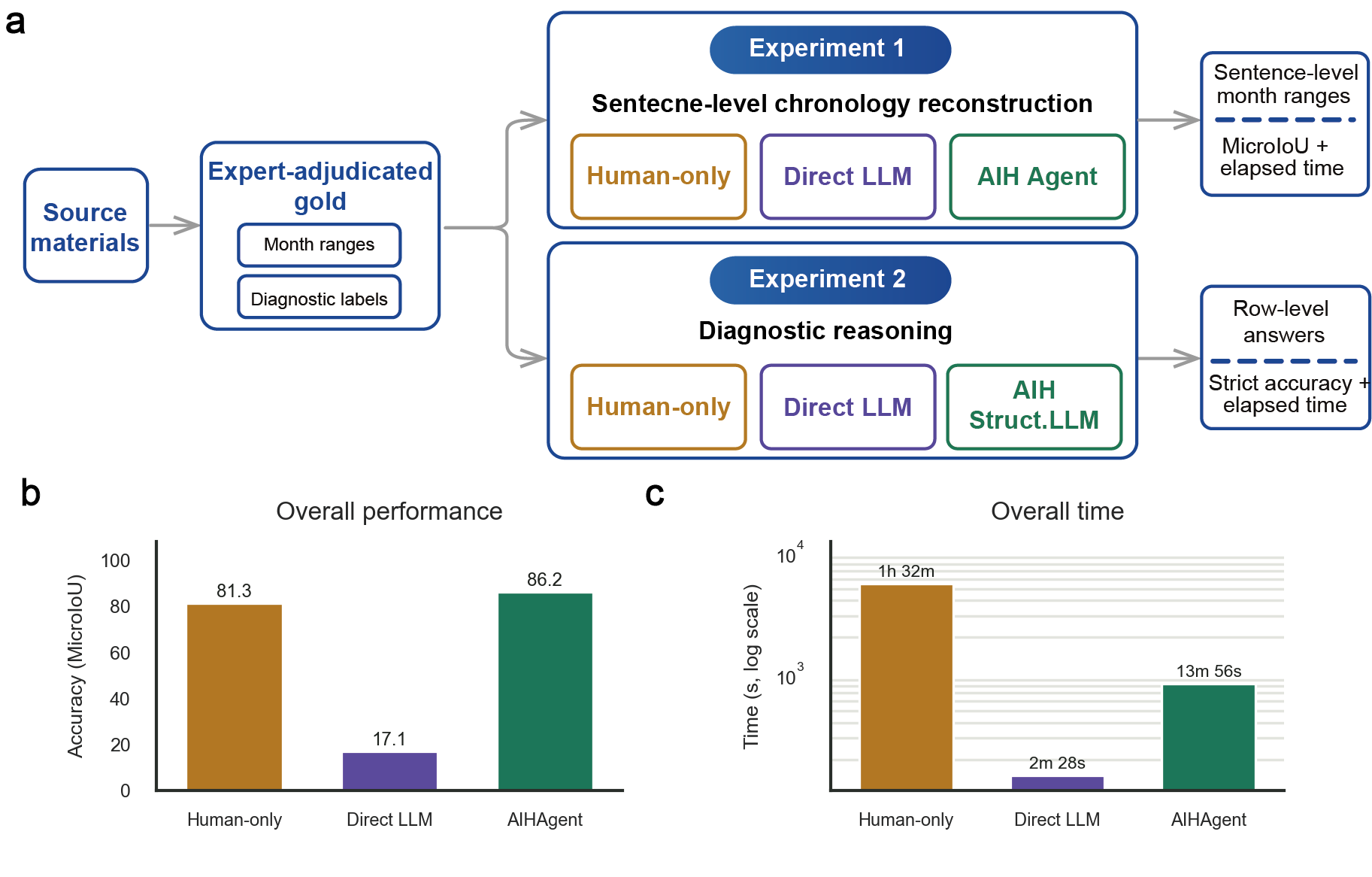}
  \caption[Evaluation design and end-to-end reconstruction of person-centred temporal ranges]{Evaluation design and end-to-end reconstruction of person-centred temporal ranges. \textbf{a,} Evaluation design: source materials were expert-annotated and adjudicated to produce sentence-level month ranges and diagnostic labels, and then entered two complementary branches. Experiment 1 compares human-only annotation, direct large-language-model prompting and AIH Agent in sentence-level temporal-range reconstruction; it outputs sentence-level month ranges and evaluates them using MicroIoU and cumulative elapsed time. Experiment 2 compares human responses, direct large-language-model prompting and structured large-language-model prompting in diagnostic reasoning; it outputs row-level answers and evaluates them using strict accuracy and cumulative elapsed time. \textbf{b,} End-to-end temporal-reconstruction performance across six evaluation cases. Bars show month-level, case-size-weighted MicroIoU for human-only annotation, direct large-language-model prompting and AIH Agent over 244 eligible sentence-level temporal ranges. \textbf{c,} Cumulative elapsed time for the same end-to-end task, shown in seconds on a logarithmic axis. Accuracy and elapsed time by case group are provided in \supfigref{supp-fig:supp-experiment1-case-groups}; accuracy and elapsed time for the diagnostic experiment are provided in \supfigref{supp-fig:supp-diagnostic-accuracy} and \supfigref{supp-fig:supp-diagnostic-time}.}
  \label{fig:evaluation-results}
\end{figure}

MicroIoU measures the overlap between predicted and reference temporal ranges at month resolution; higher values indicate more accurate temporal localization.
Across 244 eligible sentence-level temporal ranges, AIH Agent achieved a case-size-weighted MicroIoU of 86.2\%, exceeding human-only annotation (81.3\%) and direct prompting (17.1\%; \figref[b]{fig:evaluation-results}).
AIH Agent assigned temporal ranges to these sentences in 13\,min 56\,s, whereas human-only annotation required 1\,h 32\,min, meaning that AIH Agent used approximately 15\% of the human time (\figref[c]{fig:evaluation-results}).
Direct prompting was faster (2\,min 28\,s), but its low overlap with the reference ranges shows that speed alone is insufficient for reliable reconstruction of person--time evidence.

Across the six \textit{Shiji} cases, AIH reconstructed person-centred temporal ranges more accurately in less than one-sixth of the human time, with model-call costs below US\$1.

When grouped by task type, AIH Agent achieved 100.0\% MicroIoU in both direct temporal localization and ambiguous within-document reasoning. Its score was 71.2\% in cross-document propagation cases C5 and C6, but remained above human-only annotation (67.1\%) and direct large-language-model prompting (11.1\%; \supfigref{supp-fig:supp-experiment1-case-groups}; \suppsecref{supp-sec:supp-experiment1-case-group-results}). To test whether this result depended on a particular underlying model, we additionally replicated Experiment 1 using DeepSeek-V4-Flash in non-thinking mode, GPT-5.6 Sol, Gemini 3.1 Pro, Claude Opus 5 and Qwen 3.6. Across all five models, the six-case mean MicroIoU of AIH Agent was substantially higher than that of direct prompting with the same model, indicating that the principal performance gain arose from the structured workflow rather than from a particular model. Claude Opus 5 achieved the highest mean (90.7\%), followed by DeepSeek-V4-Flash (90.2\%); the latter was the only model to meet or exceed the human-only baseline in all six cases. Complete case-group, case-level and observed-time results are reported in the Supplementary Information (\supfigref{supp-fig:supp-experiment1-case-groups}; \supfigref{supp-fig:supp-experiment1-case-level}).

The diagnostic experiment produced a consistent result: compared with direct prompting, structured prompting increased accuracy in cross-text event verification and temporal alignment from 75.0\% to 91.7\%, a larger improvement than for temporal-information extraction and judgement, whereas it produced no improvement in within-document temporal reasoning (\supfigref{supp-fig:supp-diagnostic-accuracy}; \supfigref{supp-fig:supp-diagnostic-time}; \suppsecref{supp-sec:supp-diagnostic-results}).
Case analyses of C5 and C6 further illustrate this difficulty. AIH must first determine whether passages from different biographies narrate the same event or historical phase, and then use more explicit temporal information from one source to constrain a less explicit temporal range in another. The candidate cross-text relation and temporal-boundary judgement therefore form the decisive links in the final result (\suppsecref{supp-sec:supp-cross-textual-propagation-case}).
This pattern is consistent with the layered design of AIH Agent and suggests that structured decomposition and intermediate evidence representations may be important for improving cross-text judgements. Sentence-level agents first preserve locally verifiable evidence; TimeBlocks then represent temporal scope, ordering and cross-text constraints. The model therefore need not perform evidence extraction, event consolidation and temporal normalization simultaneously in a single prompt, and each judgement can be inspected and revised separately.
Overall, the end-to-end evaluation and diagnostic experiment jointly show that AIH Agent can improve temporal-localization accuracy and reduce task completion time while preserving evidence traceability. Cross-text event verification and temporal-boundary judgement remain the key stages for further performance improvements.

\subsection{Westlake Historian transforms person--time evidence into open, collaborative research infrastructure}

\begin{figure}[H]
  \centering
  \includegraphics[width=1\textwidth]{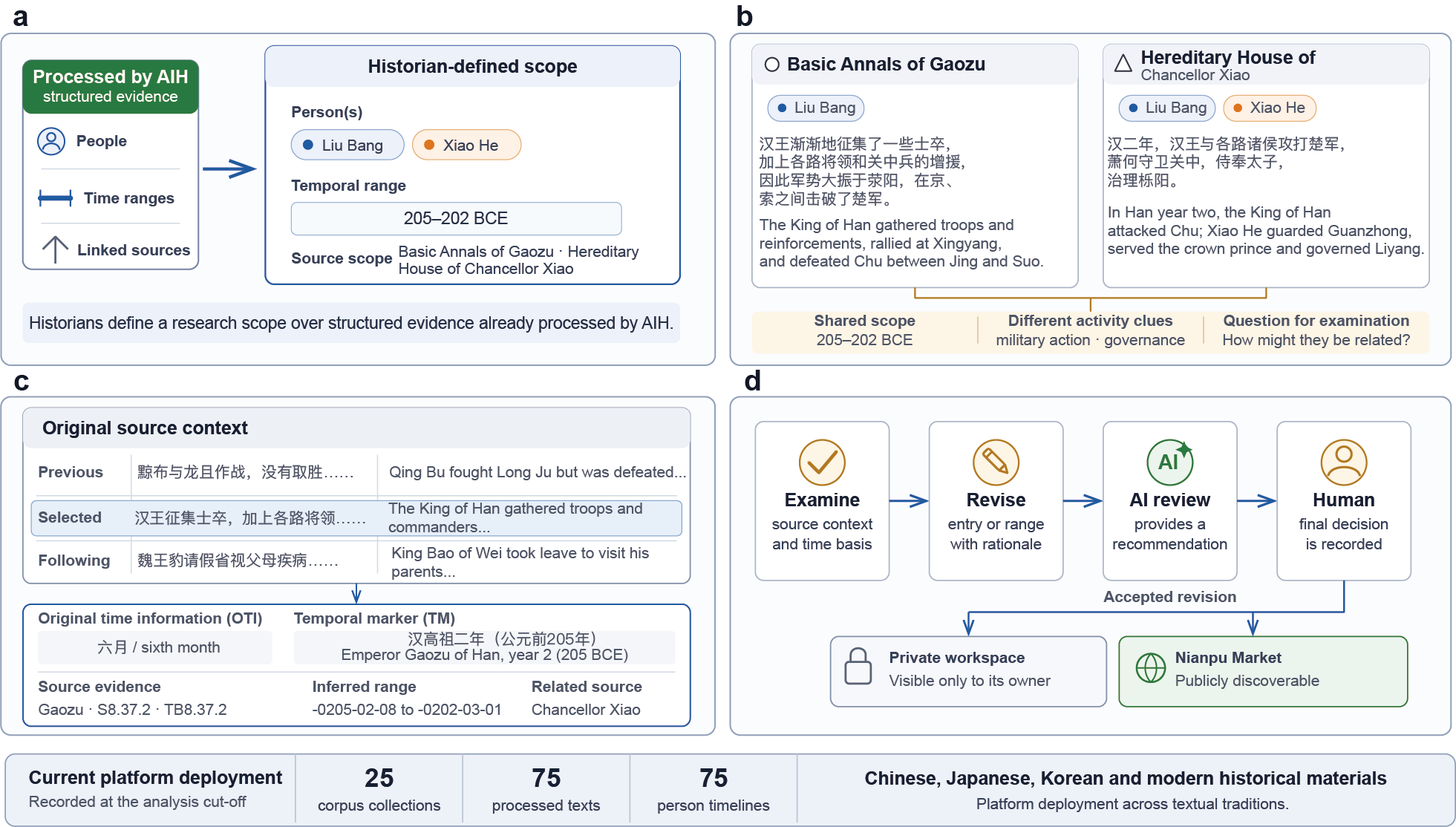}
  \caption[Historians use Westlake Historian to inspect and revise person--time evidence]{Westlake Historian incorporates AIH-processed person--time evidence into a historian-led research process. \textbf{a,} Historians begin with person records, temporal ranges and source links organized by AIH, and delimit the people, period and sources under examination. \textbf{b,} Within the same scope, the platform juxtaposes relevant original passages from different texts, allowing researchers to compare people's activities, formulate questions for investigation and use the original text to assess further the historical relations among the materials. \textbf{c,} For a selected lead, the platform presents its surrounding context, original temporal information, temporal marker completed from context, inferred temporal range and related sources, enabling the temporal judgement to be verified against the original text. \textbf{d,} Historians inspect the evidence and propose revisions to entries or temporal ranges; AI review provides suggestions, but the final decision is made and recorded by a human. Accepted versions may remain in a private workspace or be published to the Chronology Market. The numbers along the bottom summarize the processed corpus analysed in this study.}
  \label{fig:aih-platform}
\end{figure}

Automatic reconstruction is only the starting point for bringing person--time evidence into historical research. To support researchers in further inspecting, revising and sharing this evidence, we developed Westlake Historian, an open research platform based on AIH (\url{https://westlakehistorian.com}; \figref{fig:aih-platform}). The platform incorporates person records, temporal ranges and source evidence organized by AIH into a research process of delimiting scope, comparing materials, returning to the original text for verification and making human revisions. Every temporal judgement and subsequent modification can therefore be traced back to the corresponding sentence, the basis for temporal inference and related sources (Fig.~\ref{fig:aih-platform}).

At the evidence-inspection layer, historians first specify a person, temporal range and sources, after which the platform retrieves and juxtaposes relevant sentences from different texts. Figure~\ref{fig:aih-platform}a--c uses the \textit{Basic Annals of Gaozu} and the \textit{Hereditary House of Chancellor Xiao} as an example. Researchers can compare the activities described in the two biographies during the same period. After selecting a sentence, they can inspect its surrounding context, original temporal expression, temporal marker completed from context, inferred temporal range and source link, thereby verifying why the sentence was assigned to that temporal position. The platform thus supports historians in assembling materials, verifying the basis of temporal judgements and using the original texts to examine the historical relations among sources.

At the revision layer, historians can revise an entry or temporal range on the basis of the original text and temporal evidence, and provide a rationale. AI review supplies supporting suggestions, but the final decision on acceptance is made and recorded by a human (Fig.~\ref{fig:aih-platform}d). Accepted versions may remain in a private workspace or be published to the Chronology Market, where other researchers can discover and continue to examine them. This design preserves person--time evidence, original sources, the basis of temporal judgements, rationales for revision and human decisions in the same traceable object. Historians can inspect and modify each item along the evidence chain, while public versions provide common objects for textual criticism and scholarly exchange. The platform's functions and interfaces are shown in \supfigrangeref{supp-fig:supp-aih-interface}{supp-fig:supp-platform-coverage}.

The Westlake Historian collections analysed here comprise 75 texts and 75 person-centred chronology tracks spanning Chinese, Japanese and Korean histories and modern historical materials; their composition is reported along the bottom of Fig.~\ref{fig:aih-platform} and in \suptabref{supp-tab:supp-platform-coverage}. Among them, six \textit{Shiji} cases provide the evaluation of accuracy and efficiency, while the cross-regional example collections demonstrate the feasibility of using a common person--time evidence representation to organize materials from different textual traditions.

The significance of these collections lies not only in their scale, but also in whether placing dispersed accounts on a common timeline can bring historical questions obscured by chapter-based narration back into view. The retreat from Pengcheng in this project's \textit{Shiji} cases provides one example.

After the catastrophic defeat at Pengcheng, Liu Bang escaped with only a few dozen horsemen. How could a force on the verge of disintegration recover its fighting strength at Xingyang so soon afterwards? No single chapter of the \textit{Shiji} preserves a complete answer. In the vernacular materials used here, the relevant clues are distributed among the \textit{Basic Annals of Xiang Yu}, the \textit{Basic Annals of Gaozu} and the \textit{Hereditary House of Chancellor Xiao}. The \textit{Basic Annals of Xiang Yu} records that Liu Bang ``gradually gathered some of the scattered soldiers'' during his retreat. The defeated forces then converged at Xingyang, where Xiao He mobilized even the elderly and under-age people of Guanzhong who had not been registered for regular service, allowing the Han army to recover its strength. The \textit{Basic Annals of Gaozu} gives a more concise account: Liu Bang gathered soldiers, received reinforcements from other commanders and troops from Guanzhong, and subsequently defeated the Chu army between Jing and Suo. The \textit{Hereditary House of Chancellor Xiao} shifts the view to the Guanzhong rear, where Xiao He administered household registers, transported provisions and repeatedly mobilized soldiers to replace losses when Liu Bang's forces retreated (\textit{Shiji}, juan 7, \textit{Basic Annals of Xiang Yu}; juan 8, \textit{Basic Annals of Gaozu}; juan 53, \textit{Hereditary House of Chancellor Xiao}).

By reorganizing records dispersed across chapters devoted to different people, AIH brings into view a historical picture fragmented by biographical narration: after the collapse at Pengcheng, scattered troops were gathered at the front and defeated forces converged at Xingyang, while soldiers and provisions continued to arrive from Guanzhong, enabling the Han army to recover and return to battle at Jing and Suo. This juxtaposition connects front-line reassembly with mobilization in the Guanzhong rear and places a question before the historian that merits renewed examination: how did these two processes jointly shape the Han recovery from Pengcheng and the subsequent return to battle at Jing and Suo?

\section{Discussion}

Large language models and agents are transforming AI from one-shot generation tools into research collaborators capable of collecting materials in sequential steps, proposing candidate interpretations and undergoing programmatic or human evaluation. This trend is already evident in scientific hypothesis generation, chemical reasoning and experimental planning, mathematical construction and algorithm search\cite{wang2024scimon,yang2024opendomain,yang2025moosechem,bran2024chemcrow,boiko2023autonomous,romeraparedes2024funsearch,novikov2025alphaevolve,gottweis2026coscientist}. Although these studies address different disciplines, together they reflect a shift in research practice from models generating final answers to models participating in connected stages of material collection, candidate generation, verification and revision. AIH realizes this shift in historical research: AI retrieves materials at scale, organizes temporal cues and generates candidate associations, while preserving sources and intermediate judgements for evidence verification and historical interpretation by historians.

Digital history has developed historical knowledge graphs, biographical databases and social networks that structure people, places, events and relationships\cite{chen2022cbdb,hyvonen2019biographysampo,hyvonen2016warsampo}. Related work has also addressed historical date recognition, event ordering and relative timelines, temporal reasoning by language models and event knowledge graphs\cite{strotgen2014extending,leeuwenberg2018temporal,chambers2014dense,ning2018multiaxis,wang2024tram,guan2023eventkg}. AIH is related to these approaches in likewise transforming historical materials into comparable structured objects. It differs in beginning with source sentences from fragmentary historical texts, explicitly preserving the basis of temporal judgements, candidate cross-text associations and inferred ranges, and organizing people's activities across biographies on a shared timeline that remains verifiable against the original text.

Reconstructing the temporal sequence of people's activities is essential to understanding the contexts of their actions, their interactions and broader historical processes. Yet accounts of the same person or event are often dispersed across texts, chapters and narrative perspectives. Researchers must determine not only which people and events the materials concern, but also when they occurred, how they are related and how accounts from different sources compare.

This work poses two interrelated challenges. First, temporal judgements may derive from explicit dates in the original text, contextual completion or constraints supplied by other chapters, making the evidence and inferential process difficult to preserve systematically. Second, material retrieval, temporal ordering and cross-text comparison require substantial human labour. According to the publisher's account of the preliminary \textit{Extended Chronological Biography of Hu Shi}, for example, the overall project lasted 17 years, including approximately five years of active compilation\cite{lienching1984hushi}.

Why does direct prompting fail so readily in this setting? It asks a model to identify people and time, reorder retrospective passages, merge events across chapters and normalize dates in a single response. An early error can alter the entire timeline, while the researcher is often left with only a final date that cannot be interrogated. AIH changes this arrangement. It first preserves source sentences and then uses TimeBlocks to mark which sentences share a temporal context. Narrative and chronological order are handled separately, and a date from another chapter can constrain a passage only after the relationship has been verified against the source text. The final date is therefore no longer a model conclusion detached from the record, but part of an evidence chain that can be followed back to the original passages. A historian can accept one judgement, revise another or reject an association with insufficient evidence without accepting or discarding the model's answer as a whole. The value of the structured workflow lies not in generating more conclusions, but in giving each conclusion a provenance that can be questioned and revised. Researchers can thus compare the sequence and simultaneity of people's actions, identify cross-text leads worth pursuing, formulate questions and return to the original texts for source criticism.

Across the six evaluated \textit{Shiji} case packets, AIH Agent reconstructed temporal ranges more accurately than human-only annotation and direct large-language-model prompting, while reducing task completion time (\figref[b,c]{fig:evaluation-results}). Structured prompting produced its largest gain in cross-text event verification and temporal alignment (\supfigref[b]{supp-fig:supp-diagnostic-accuracy}), indicating that separating temporal-cue identification, adjustment of narrative order, cross-text evidence verification and temporal-range normalization can improve performance on this task. The cross-document cases further show that determining whether candidate texts describe the same event, whether person references are consistent and where narrative boundaries fall should be priorities for future improvement.

The Pengcheng case in the Results illustrates that the historiographical value of AIH lies not in turning juxtaposition into a causal claim, but in changing the field of materials that a researcher can inspect at once. Clues from the front and the rear that were assigned to different biographies can be examined together and then pursued as a question back into each source chapter. AIH-generated chronology thus not only reconstructs the sequence of people's activities, but also turns connections that may have been obscured by chapter-based narration into testable historical leads.

From a historiographical perspective, temporarily rearranging materials from the annals-biographies format into an annalistic view does not replace one form with another. The biographical form preserves individual perspectives, narrative selection and structures of praise and blame; an annalistic view facilitates comparison among actions occurring at the same time. Moving between the two allows researchers both to identify leads across chapters and to return to the original chapter to assess context and relationships among sources. AIH therefore serves as an inspectable material layer for extended biographical chronologies, textual criticism and cross-text comparison. It does not write definitive conclusions for historians, but reorganizes materials accumulated over long periods and entangled across texts into evidence that can be compared and verified, allowing new historical questions to emerge.

As materials from more regions, periods and languages are incorporated, previously dispersed records of people and events can be juxtaposed and compared at a broader scale, helping historians identify research leads across texts, people and even historiographical traditions. AI Historian thus represents a future paradigm for human--AI collaborative historical research. While keeping every judgement traceable to primary sources, AI can automatically retrieve, organize and associate evidence across large corpora and propose candidate associations linked to their original texts. Historians can then use source context to judge whether candidate materials are related, reject associations with insufficient evidence, revise temporal boundaries and deepen interpretation, enabling historical research to formulate and test new questions and interpretations continuously at a larger evidentiary scale.

In summary, across 244 sentence-level temporal ranges in six \textit{Shiji} cases, AIH Agent achieved a MicroIoU of 86.2\% in a cumulative 13\,min 56\,s, whereas human-only annotation required 1\,h 32\,min. This quantitative evaluation, together with the cross-regional deployment of Westlake Historian, indicates that AIH can improve the accuracy of person--time evidence reconstruction while reducing human effort, and can support the organization and verification of materials from different textual traditions through a common, traceable and revisable representation. Its significance lies not in replacing historians' judgements about sources, but in combining large-scale material processing with human interpretation in an inspectable human--AI workflow.

\section{Methods}

\subsection{Evaluation-case selection and input texts}

The end-to-end evaluation in Experiment 1 uses six predefined case packets (C1--C6) extracted from modern Chinese vernacular translations of three \textit{Shiji} chapters closely related to the Chu--Han Contention and the formation of the early Han political order: the \textit{Basic Annals of Xiang Yu}, the \textit{Basic Annals of Gaozu} and the \textit{Hereditary House of Chancellor Xiao}. The chapters provide the perspectives of Xiang Yu, Liu Bang's principal rival in the Chu--Han Contention; Liu Bang, founder of the Han dynasty; and Xiao He, a key administrator and logistics organizer in Liu Bang's camp. C1--C2 select direct temporal-localization passages from the biographies of Xiao He and Xiang Yu, respectively; C3--C4 select ambiguous within-document passages from the biographies of Liu Bang and Xiang Yu, respectively; and C5--C6 pair cross-text passages from the \textit{Basic Annals of Gaozu} with the \textit{Basic Annals of Xiang Yu} and the \textit{Hereditary House of Chancellor Xiao}, respectively.
The three biographies provide interrelated source contexts with different perspectives for case selection and cross-text comparison. The six case packets contain 244 eligible sentence-level temporal ranges. Detailed case sources are provided in \suppsecref{supp-sec:supp-corpus-scope}.
The evaluation data associate source sentences, person annotations and expert-adjudicated reference temporal ranges in each case with a unified timeline for end-to-end temporal reconstruction and scoring.

Experiment 2 uses a separate question set drawn from the same three vernacular \textit{Shiji} translations. It decomposes the AIH workflow into three diagnostic forms (T1--T3), each with 16 questions: eight on temporal-information extraction and judgement, four on temporal-information reasoning and four on cross-text event verification and temporal alignment. Together, the three forms comprise 48 unique questions; each form was completed by two participants, yielding 96 scored human responses. Under the model conditions, direct and structured prompting were compared on the same 48 questions and evaluated by strict row accuracy and cumulative elapsed time. The question composition and participant allocation for each form are provided in \supfigref{supp-fig:supp-participant-assignment}.

AIH is an agent system built on general-purpose large language models. In the case of the \textit{Shiji}, Classical Chinese materials require additional word segmentation, part-of-speech tagging and normalization of personal names before automated analysis. Existing tools achieve F-scores of approximately 85--90\% on ancient Chinese text and still require human checking, while aliases and distinct people sharing the same name further complicate entity disambiguation\cite{li2020shiji}. For the present task, ellipsis and relative dating introduce additional temporal-reasoning difficulty. Benchmark studies also show that large language models perform substantially worse on Classical Chinese than on modern Chinese, with particular instability in semantic inference, historical knowledge and long-text comprehension, and marked variation across models\cite{zhang2023aclue,wei2024aceval}. Direct use of Classical Chinese would therefore cause an evaluation to reflect both the chronology workflow and the underlying model's ability to understand Classical Chinese, making their contributions difficult to separate. Both experiments consequently use modern Chinese vernacular translations of the \textit{Shiji}, focusing evaluation on temporal-cue identification, TimeBlock construction, temporal normalization, cross-text event verification and temporal alignment rather than model-specific competence in Classical Chinese.
Experiment 1 tests end-to-end temporal reconstruction, whereas Experiment 2 tests its key component judgements.

\subsection{AIH Agent workflow and TimeBlock representation}

Event-centred knowledge graphs can model temporal relations and generate biographical timelines\cite{gottschalk2019eventkg}, and existing historical networks are effective at revealing relationships among people\cite{warren2016sixdegrees,bornhofen2020dynamic}. Reconstructing traceable person-centred timelines from fragmented source texts, however, also requires explaining when actions occurred, where the evidence comes from and how dispersed evidence was connected. Relative dates, ellipsis, retrospective passages and cross-chapter references in histories in the annals-biographies format further increase this difficulty\cite{strotgen2014extending,leeuwenberg2018temporal}; cross-sentence event ordering and the distinction between temporal axes also present problems of accuracy and annotation ambiguity\cite{chambers2014dense,ning2018multiaxis}.
AIH operationalizes these requirements through sentence-level identification of temporal cues and TimeBlock ordering. Extending historical knowledge graphs that rely primarily on structured or semi-structured materials\cite{chen2022cbdb,hyvonen2019biographysampo,hyvonen2016warsampo}, AIH organizes and verifies cross-chapter temporal evidence from fragmented biographical texts while preserving provenance. It decomposes identification, ordering, verification and temporal-range generation into inspectable steps to address the temporal-reasoning demands documented for direct prompting\cite{wang2024tram}.

AIH Agent implements nine semantic agents in an auditable chronology-construction workflow (\figref{fig:aih-multi-agent-pipeline}a).
The agents operate at two representational levels.
A1--A4 operate at the sentence level: they identify people in sentences, extract temporal expressions from the original text, determine the narrative function of each sentence and mark shifts in narrative time caused by retrospective passages, converting continuous text into structured signals for TimeBlock construction.
A5--A9 operate at the TimeBlock level: A5--A7 complete and order TimeBlocks within a document, A8 verifies cross-biography evidence, and A9 converts anchors and verified evidence into normalized internal temporal coordinates.
Formal temporal prediction ends at A9.
Non-agent preprocessing, initial TimeBlock assembly and generation of the final human-readable summary are pipeline stages rather than semantic agents and are therefore omitted from the main figure; the expanded architecture is provided in \supfigref{supp-fig:supp-aih-agent-architecture}, and the functions and outputs of all nine semantic agents in \suptabref{supp-tab:supp-nine-agent}.

AIH represents each biography at two linked levels: sentence-level signals and TimeBlock-level chronological units.
At the sentence level, the workflow preserves person annotations, temporal cues, text-type labels and markers of retrospective passages so that later chronological claims remain connected to source passages.
At the TimeBlock level, adjacent sentences governed by the same temporal marker are grouped; underspecified temporal markers are completed from context; and TimeBlocks containing retrospective passages can be reordered when narrative order differs from chronological order.
Cross-text propagation likewise operates on the TimeBlock representation.
A8 records candidate relations and citation-level evidence between source and target TimeBlocks, and marks whether each relation meets the propagation criteria; A9 converts only evidence-verified relations into normalized internal anchors or boundary constraints.
Thus, a final temporal range is not a copied date, but the result of within-text order, candidate anchors and cross-text constraints acting together in the target chronology.
Detailed sentence fields and TimeBlock variables are provided in \suptabref{supp-tab:supp-sentence-fields} and \suptabref{supp-tab:supp-timeblock-fields}; construction and cross-text propagation are illustrated in \supfigref{supp-fig:supp-timeblock-construction} and \supfigref{supp-fig:supp-cross-textual-propagation}.

Experiment 2 does not treat questionnaire responses as outputs of complete AIH Agent. Instead, it separately tests intermediate judgements central to the workflow: temporal-information extraction and judgement corresponds to sentence-level signal identification; temporal-information reasoning corresponds to temporal-marker completion and TimeBlock ordering; and cross-text event verification and temporal alignment corresponds to cross-text constraint judgements grounded in source evidence. This decomposition allows diagnostic results to identify which reasoning stage may contribute to end-to-end performance differences rather than combining all errors into a single aggregate score.

\subsection{Human and model evaluation design}

We conducted two complementary evaluations of structured temporal reasoning (Table~\ref{tab:evaluation-design}).
Experiment 1 evaluates end-to-end sentence-level temporal reconstruction over six case packets.
Predicted and reference temporal ranges were normalized to month resolution and compared using MicroIoU.
For each eligible sentence \(s\), after open boundaries were clipped to the evaluation window for that case, reference and predicted temporal ranges were projected to finite month sets \(G_s\) and \(P_s\), respectively.
MicroIoU is calculated as:
\[
\mathrm{MicroIoU}=\frac{\sum_s |G_s \cap P_s|}{\sum_s |G_s \cup P_s|}.
\]
Sentences without a gold-standard temporal range do not enter either the numerator or denominator; for sentences with a known gold-standard range but no prediction, the predicted set is empty.
The overall value reported in \figref[b]{fig:evaluation-results} is the case-size-weighted mean of case-level MicroIoU over 244 eligible sentence-level temporal ranges.
Human-only results come from six participants, each of whom independently completed one assigned case without AI assistance.
AIH Agent results were formed through row-level consensus across three independent runs, using majority agreement and an interval medoid to resolve ties without access to gold-standard answers.

The six cases in Experiment 1 contain 249 source sentences in vernacular Chinese (8,236 Chinese characters, including punctuation) drawn from the \textit{Basic Annals of Gaozu}, the \textit{Basic Annals of Xiang Yu} and the \textit{Hereditary House of Chancellor Xiao}, and centre on Liu Bang, Xiang Yu and Xiao He. Of these sentences, 244 sentence-level temporal ranges entered the evaluation. The reported 13\,min 56\,s is the cumulative time for three independent runs and generation of their consensus, averaging approximately 2\,min 19\,s per case; when expressed across the three core person-specific tracks, this corresponds to approximately 4\,min 39\,s per track. A conservative conversion based on the input scale and public API prices places the model-call cost for Experiment 1 below US\$1.

Experiment 2 evaluates component judgements required by the workflow with strict row accuracy: an item is counted as correct only when every required field matches the reference answer.
The human condition comprises 96 scored responses from six completed forms; direct and structured large-language-model prompting each comprise the same 48 unique questions.
Model results were determined by majority vote across three independent runs.
For both experiments, elapsed time within each displayed condition is cumulative task time.

\begin{table}[H]
\centering
\caption{Design and reporting criteria for the human and model evaluations.}
\label{tab:evaluation-design}
\small
{\renewcommand{\tabularxcolumn}[1]{m{#1}}%
\begin{tabularx}{\textwidth}{@{}M{0.12\textwidth}M{0.20\textwidth}M{0.22\textwidth}Y M{0.17\textwidth}@{}}
\toprule
Evaluation & Conditions & Evaluation set & Primary metric & Aggregation and timing \\
\midrule
Experiment 1: temporal reconstruction &
Direct large-language-model prompting; human-only; AIH Agent &
Six case packets; 244 eligible sentence-level temporal ranges &
Month-level MicroIoU between predicted and reference ranges &
Case-size-weighted mean; six human participants; AIH Agent row-level consensus across three runs; cumulative elapsed time \\
\addlinespace
Experiment 2: diagnostic reasoning &
Human; direct large-language-model prompting; structured large-language-model prompting &
96 human responses across six forms; 48 unique questions per model condition &
Strict row accuracy; all required fields must match &
Majority vote across three model runs; total elapsed time by condition \\
\bottomrule
\end{tabularx}
}
\end{table}

\subsection{Implementation and reproducibility}

The AIH pipeline was implemented as a staged workflow that separates text preprocessing, sentence-level annotation, TimeBlock reasoning, cross-text propagation, internal temporal-range normalization and final summary generation.
Intermediate outputs are saved at every rule-based and model-based stage.
Researchers can therefore trace a final chronology entry back through its normalized range, TimeBlock, sentence identifier and original source text.
In Experiment 1, AIH Agent generated predictions without access to gold-standard ranges or participant outputs; the gold standard and human annotations were read by the scoring procedure only after the consensus prediction table had been finalized.
In Experiment 2, reference answers were held in a separate scoring table and were not provided to human or model conditions; each condition received only the same question materials and its corresponding prompts, and responses were compared field by field after completion.
The aggregation rule for the three model runs and the timing convention are described in the preceding subsection and \suppsecref{supp-sec:supp-reproducibility}.
Elapsed time was recorded as cumulative task time for each displayed condition.
The agent workflow and data representation are detailed in \suppsecref{supp-sec:supp-methods}; representative final TimeBlock outputs are provided in \suppsecref{supp-sec:supp-structured-output}, \suptabref{supp-tab:supp-timeblock-overview} and \suptabref{supp-tab:supp-final-excerpts}. Terminology, source-title renderings and historical names used across the English and Chinese versions are listed in \suptabref{supp-tab:supp-terminology}, \suptabref{supp-tab:supp-text-titles} and \suptabref{supp-tab:supp-person-names}.

Material scale was calculated directly from the input files and structured outputs of the 25 collections. Character counts include punctuation; people were deduplicated using the person labels in the files; and sentences, TimeBlocks and temporal anchors were counted from the corresponding output objects. Runtime was summed from the wall-clock time in the latest successfully completed log for each collection and pipeline stage, yielding 205\,h in total. Under the current four-way concurrency setting of the model endpoint, the estimated processing time is approximately 51\,h (2.1\,d). On the basis of the measured time difference between AIH and human-only annotation in Experiment 1, organizing material at the same scale is estimated to require approximately 169 eight-hour working days for one researcher. API cost was estimated from the recorded invocation scale and the public list price of Qwen3.6-35B-A3B: RMB 1.8 and RMB 10.8 per million input and output tokens, respectively, yielding approximately RMB 1,000--2,000 (US\$147--295). These time and cost figures are order-of-magnitude estimates.

\section{Data availability}

The three vernacular \textit{Shiji} translations used for method validation, together with their derived annotations, ordering sequences and final TimeBlock outputs, are available through Westlake Historian at \url{https://westlakehistorian.com}. The platform supports tracing final chronology entries back to TimeBlocks, sentence positions and source-text evidence.

\section{Code availability}

The source code and accompanying materials for AIH are publicly available at \url{https://github.com/YiFengLu1999/AI-Historian}.

\section{Acknowledgements}

We thank the humanities scholars who provided valuable assistance for this study: Quanxiu Li, Xiaoxue Zhao, Jing Feng, Yichen Wang and Cong Wang from Peking University; Xiang Li from Beijing University of Chinese Medicine; Junwei Huang from Zhejiang Chinese Medical University; and Zhuofan Zhao and Yutao Zhao from Tsinghua University. This study was supported by the Westlake University Natural Language Processing Laboratory.

\section{Author contributions}

Yifeng Lu and Zijie Yang proposed the research question, designed the AIH chronology framework, organized the corpus, implemented the main codebase and wrote the manuscript.
Jie Li contributed to dataset design and annotation, evaluation design and manuscript writing.
Qingkai Min contributed to the discussion and implementation of the core algorithms.
Yue Zhang supervised the research design, methodological framing and manuscript revision.
All authors read and approved the final manuscript.

\section{Competing interests}

The authors declare no competing interests.

\bibliographystyle{unsrt}
\bibliography{refs}

\clearpage
\let\aihstoredgeometry\geometry
\let\geometry\newgeometry
\aihsetupsupplement
\let\geometry\aihstoredgeometry
\ctexset{contentsname = Contents}
\captionsetup[figure]{name=Supplementary Fig.}
\captionsetup[table]{name=Supplementary Table}
\renewcommand{\figref}[2][]{\textbf{Supplementary Fig.~\ref{#2}#1}}
\renewcommand{\tabref}[2][]{\textbf{Supplementary Table~\ref{#2}#1}}
\renewcommand{\suppsecref}[2][]{\textbf{Supplementary Section~\ref*{#2}#1}}
\renewcommand{\suptabref}[2][]{\textbf{Supplementary Table~\ref*{#2}#1}}
\renewcommand{\supfigref}[2][]{\textbf{Supplementary Fig.~\ref*{#2}#1}}
\renewcommand{\supfigrangeref}[2]{\textbf{Supplementary Figs.~\ref*{#1}--\ref*{#2}}}
\hypersetup{pdftitle={AIH Supplementary Information}}
\definecolor{AssignmentSoloFill}{HTML}{DCEBFA}
\definecolor{AssignmentSoloLine}{HTML}{5C88C5}
\definecolor{AssignmentDiagFill}{HTML}{DFF2E2}
\definecolor{AssignmentDiagLine}{HTML}{5B9B6A}
\newcommand{\assignmentbox}[3]{\fcolorbox{#1}{#2}{\rule{0pt}{1.8ex}\hspace{0.45em}\textsf{#3}\hspace{0.45em}}}
\newcommand{\solocase}[1]{\assignmentbox{AssignmentSoloLine}{AssignmentSoloFill}{#1}}
\newcommand{\diagnosticform}[1]{\assignmentbox{AssignmentDiagLine}{AssignmentDiagFill}{#1}}
\makeatletter
\newcommand{\supplementtableofcontents}{%
  \section*{\contentsname}%
  \@starttoc{spt}%
  \let\aihoriginaladdcontentsline\addcontentsline
  \renewcommand{\addcontentsline}[3]{%
    \def\aihrequestedfile{##1}%
    \def\aihtocfile{toc}%
    \ifx\aihrequestedfile\aihtocfile
      \aihoriginaladdcontentsline{spt}{##2}{##3}%
    \else
      \aihoriginaladdcontentsline{##1}{##2}{##3}%
    \fi}}
\makeatother

\setcounter{section}{0}
\setcounter{subsection}{0}
\setcounter{subsubsection}{0}
\setcounter{figure}{0}
\setcounter{table}{0}
\setcounter{secnumdepth}{3}
\section*{Supplementary Information}
\supplementtableofcontents
\newpage

\section{Biographical chronology and source organization}

\begin{figure}[H]
\centering
\includegraphics[width=0.80\textwidth]{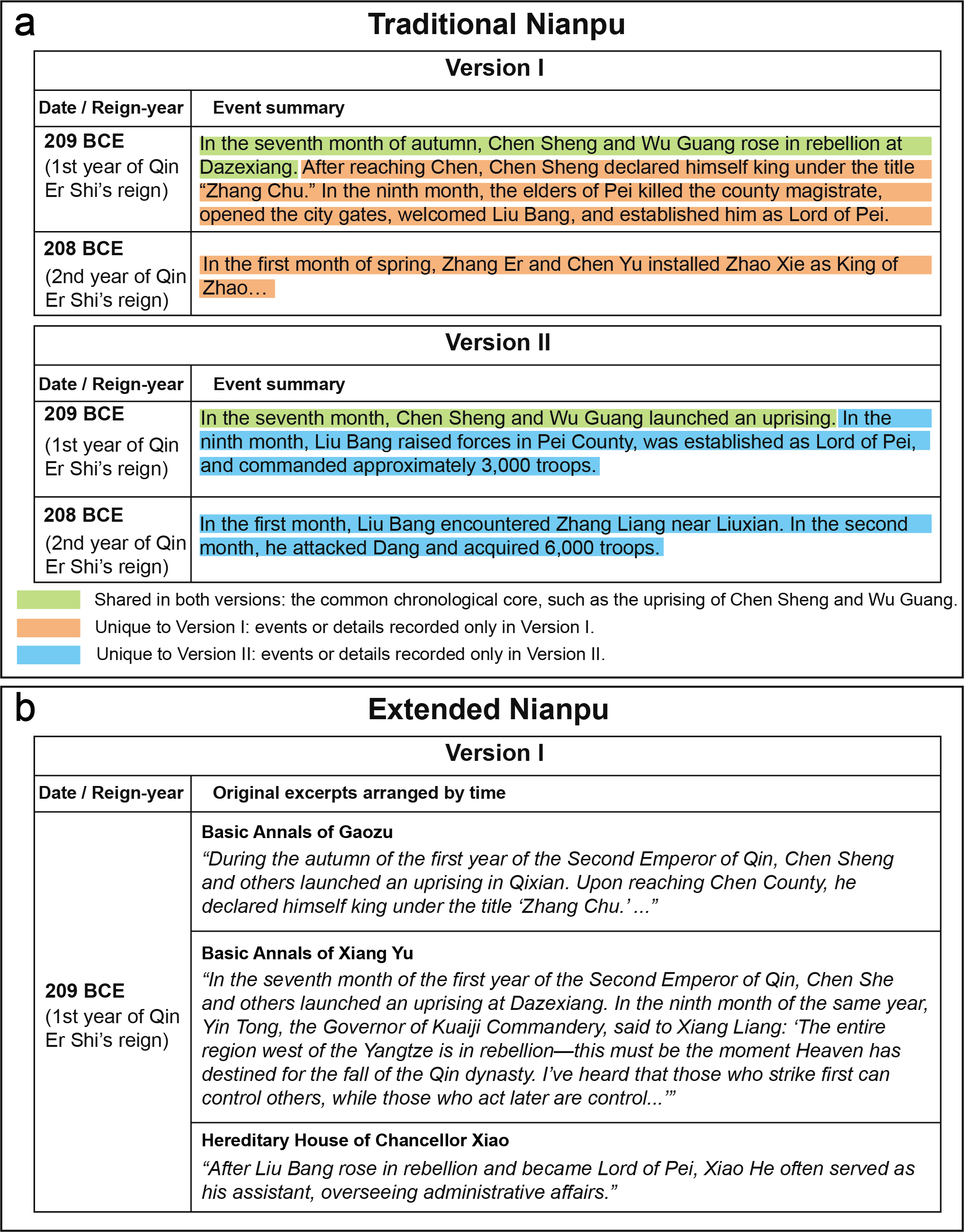}
\caption{Traditional and extended biographical chronologies. \textbf{a,} A traditional biographical chronology records the principal events of a person's life in temporal order. \textbf{b,} An extended biographical chronology preserves selected source excerpts under dated entries; source selection and quotation scope can produce multiple chronological versions for the same person.}
\label{fig:supp-nianpu-forms}
\end{figure}

\begin{figure}[H]
\centering
\includegraphics[width=0.95\textwidth]{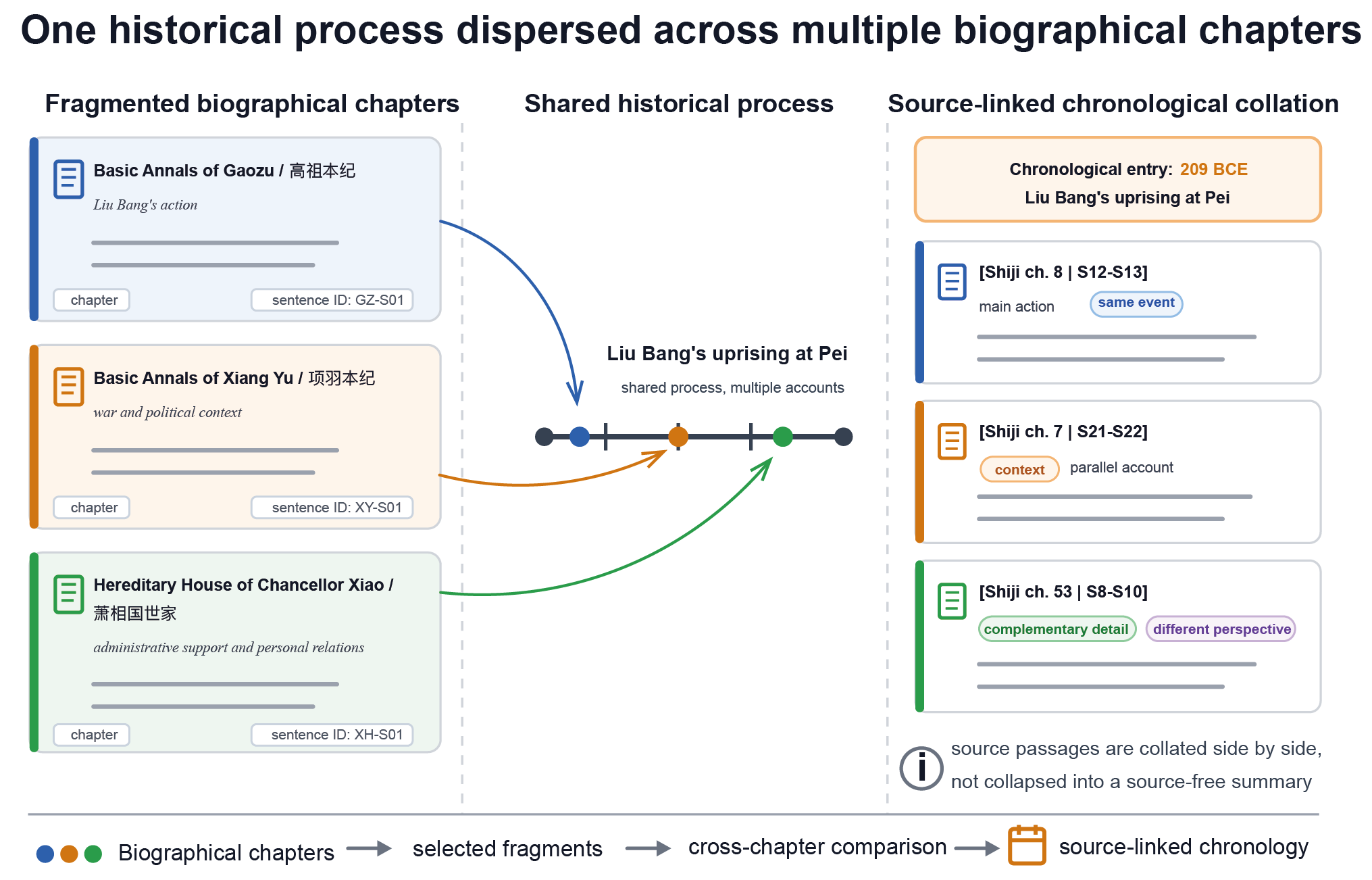}
\caption{Biographical material dispersed across chapters. A historical process may be distributed across several biographies, each preserving passages from a participant's perspective or a different narrative context. AIH compares these fragments across chapters and organizes them as source-linked chronological entries while preserving their original chapter and sentence locations.}
\label{fig:supp-biographical-fragmentation}
\end{figure}

\section{Evaluation and methods}
\label{sec:supp-methods}
\subsection{Evaluation design}
\label{sec:supp-corpus-scope}

\paragraph{Evaluation corpus and cases.}
Experiment 1 used six case packets (C1--C6) sampled in advance from modern-Chinese translations of three \textit{Shiji} chapters: the \textit{Basic Annals of Xiang Yu} (chapter 7), the \textit{Basic Annals of Gaozu} (chapter 8) and the \textit{Hereditary House of Chancellor Xiao} (chapter 53), representing the biographical perspectives of Xiang Yu, Liu Bang and Xiao He. C1--C2 were direct temporal-localization cases within a single biography, drawn from the \textit{Hereditary House of Chancellor Xiao} and the \textit{Basic Annals of Xiang Yu}; C3--C4 were within-document ambiguity-reasoning cases from the \textit{Basic Annals of Gaozu} and the \textit{Basic Annals of Xiang Yu}; C5 paired the \textit{Basic Annals of Gaozu} with the \textit{Basic Annals of Xiang Yu}, and C6 paired the \textit{Basic Annals of Gaozu} with the \textit{Hereditary House of Chancellor Xiao}, for cross-text temporal propagation. Together, the six packets contained 244 eligible sentence-level temporal ranges, which formed the scoring scope for the MicroIoU and cumulative-time results reported in the main text.

The full text of each chapter supported case selection, cross-text context and whole-chapter output inspection. The cases used modern-Chinese translations to reduce confounding from differences among models in Classical Chinese processing. AIH organized the full-chapter results at three levels: sentence-level annotations, TimeBlock representations and ordered sequences.

\begin{figure}[H]
\centering
\small
\setlength{\tabcolsep}{1.5em}
\renewcommand{\arraystretch}{1.35}
\begin{tabular}{@{}c c c@{}}
\toprule
Participant & Stage 1 & Stage 2 \\
\midrule
P1 & \solocase{C1} & \diagnosticform{T1} \\
P2 & \solocase{C2} & \diagnosticform{T2} \\
P3 & \solocase{C3} & \diagnosticform{T3} \\
P4 & \solocase{C4} & \diagnosticform{T1} \\
P5 & \solocase{C5} & \diagnosticform{T2} \\
P6 & \solocase{C6} & \diagnosticform{T3} \\
\bottomrule
\end{tabular}
\caption{Participant allocation across the two evaluation stages. Stage 1 was the end-to-end evaluation under the human-only condition, and Stage 2 was the component-level diagnostic evaluation. Blue boxes indicate cases C1--C6 and green boxes indicate diagnostic forms T1--T3. C1--C2 evaluate direct temporal localization, C3--C4 within-document ambiguity reasoning and C5--C6 cross-document temporal propagation. Each case was completed once by one participant, and each diagnostic form by two participants.}
\label{fig:supp-participant-assignment}
\end{figure}

\label{sec:supp-reproducibility}
\paragraph{Models and run settings.}
The archived Experiment 1 package used the same inputs, scoring procedure and aggregation rule for all six cases. AIH Agent was run independently three times for each model and case; results were aggregated by row-wise majority, with the medoid selected when interval outputs were tied. The direct-prompting condition for each model was run once. In Experiment 2, direct and structured prompting were each run independently three times and aggregated by row-wise majority for every question.

\label{sec:supp-human-evaluation-limitations}
\paragraph{Reference answers and human condition.}
The six participants were anonymized as P1--P6. In Experiment 1, each participant independently completed one case. In Experiment 2, each participant completed one diagnostic form, producing two responses for each of the three forms. The research team completed expert annotation and adjudication of reference answers before scoring and stored them separately from participant and model inputs. The scoring program read the reference answers after outputs from all conditions had been archived.

\subsection{AIH architecture and data representation}

\begin{figure}[H]
\centering
\includegraphics[width=0.95\textwidth]{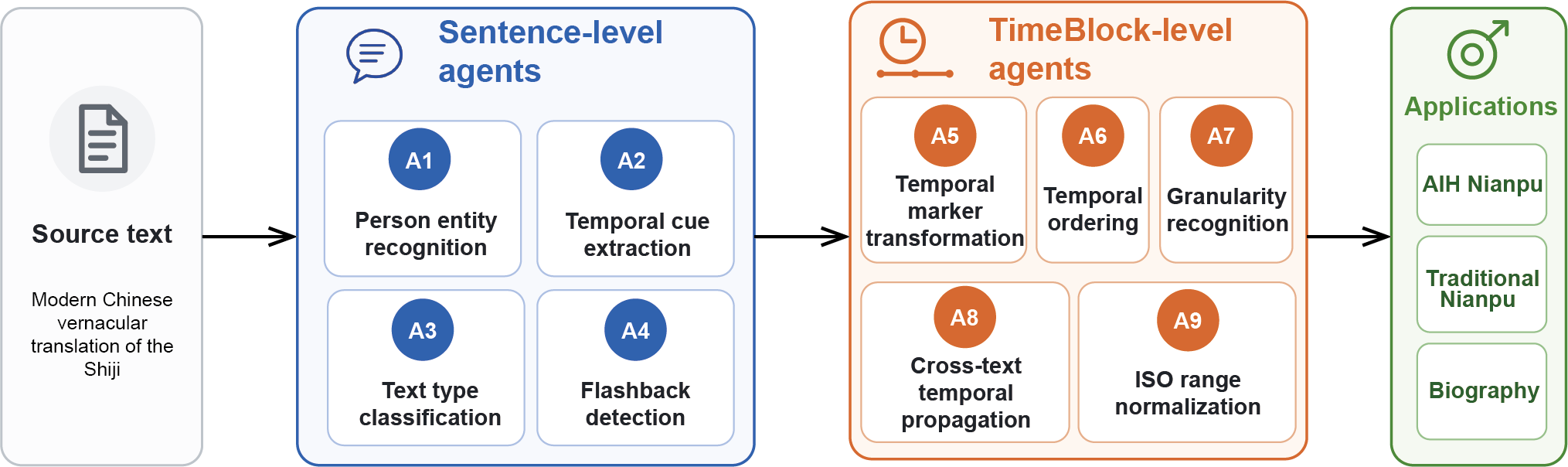}
\caption{Complete AIH Agent architecture. Sentence-level agents A1--A4 identify person entities, extract temporal cues, classify textual function and detect retrospective passages. TimeBlock-level agents A5--A7 transform temporal markers, reconstruct temporal order and determine temporal granularity. A8 coordinates cross-text temporal propagation, and A9 converts eligible anchors and constraints into normalized temporal ranges. Preprocessing, TimeBlock assembly and summary generation are supporting pipeline steps; A1--A9 are the nine semantic agents.}
\label{fig:supp-aih-agent-architecture}
\end{figure}

\begin{figure}[H]
\centering
\includegraphics[width=0.95\textwidth]{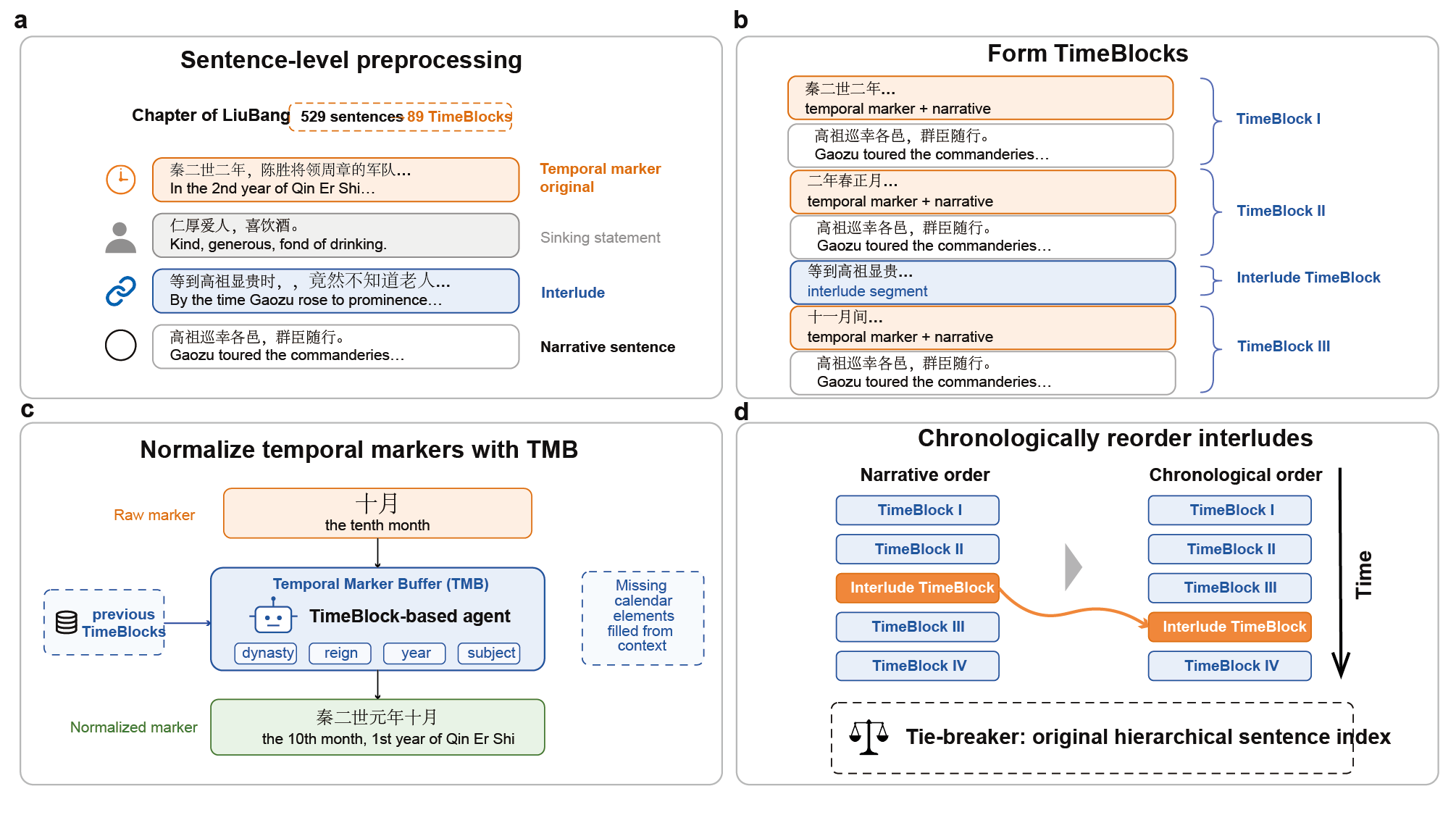}
\caption{Expanded TimeBlock construction and temporal ordering. \textbf{a,} Sentence-level preprocessing labels source temporal expressions, sinking statements, retrospective passages and ordinary narrative sentences. \textbf{b,} The labelled sequence is assembled into regular TimeBlocks and TimeBlocks containing retrospective passages. \textbf{c,} The Temporal Marker Buffer completes missing calendar elements from preceding TimeBlocks and local context. \textbf{d,} TimeBlocks containing retrospective passages are reordered from narrative into chronological order; when several blocks share a temporal anchor, the original hierarchical sentence number provides a secondary ordering key.}
\label{fig:supp-timeblock-construction}
\end{figure}

\begin{figure}[H]
\centering
\includegraphics[width=0.95\textwidth]{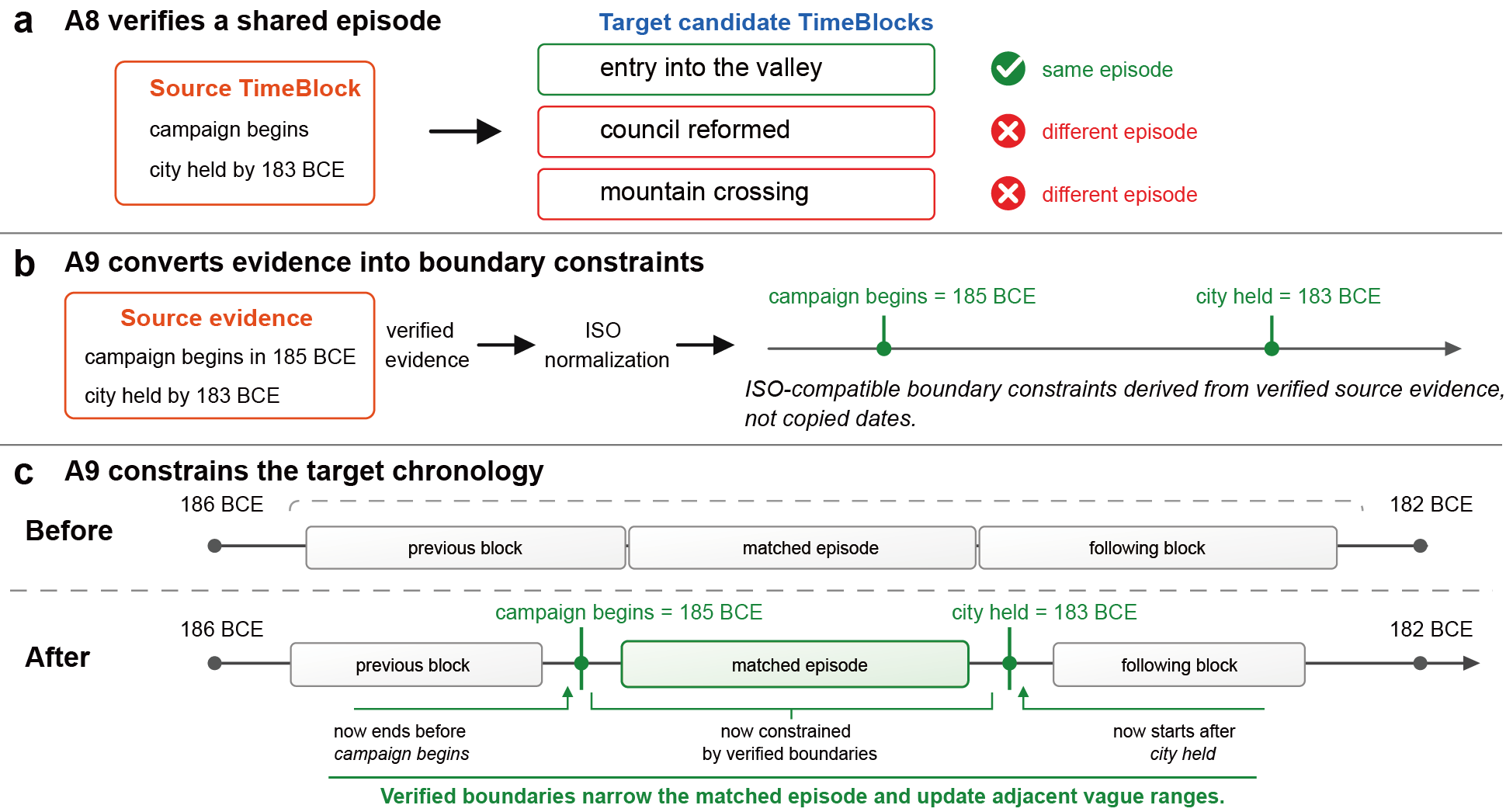}
\caption{Expanded cross-text temporal propagation. \textbf{a,} A8 compares a source TimeBlock with target candidates and retains relations supported by evidence of the same event or historical phase. \textbf{b,} A9 converts the source evidence into normalized boundary constraints. \textbf{c,} A9 applies these boundaries in the target chronology, narrowing the candidate event range and updating adjacent open temporal ranges.}
\label{fig:supp-cross-textual-propagation}
\end{figure}

\subsubsection{Agent workflow}
\label{sec:supp-nine-agent-workflow}
\paragraph{Agent responsibilities.}
AIH comprises nine semantic agents operating at sentence and TimeBlock levels. A1--A4 identify people, extract temporal expressions, classify textual function and mark retrospective passages. A5--A9 perform temporal-marker conversion, semantic ordering, granularity and anchor assessment, cross-text temporal propagation and temporal-range normalization. Text preprocessing and stable sentence numbering (step 1), initial TimeBlock assembly (step 6) and readable-summary generation (step 12) support these semantic stages.

\begin{table}[H]
\centering
\caption{Levels, functions and principal outputs of the nine AIH semantic agents.}
\label{tab:supp-nine-agent}
\small
\begin{tabularx}{\textwidth}{@{}P{0.10\textwidth}P{0.22\textwidth}P{0.40\textwidth}P{0.14\textwidth}@{}}
\toprule
Agent & Level & Principal function & Main output \\
\midrule
A1 & Sentence & Person recognition and construction of cross-text traceable referent sets. & Person fields \\
A2 & Sentence & Extraction of source temporal expressions. & Source temporal expression \\
A3 & Sentence & Text-function classification and identification of descriptive or preparatory material for sinking. & Sinking flag \\
A4 & Sentence & Detection of retrospective passages and temporal displacement from the surrounding narrative sequence. & Retrospective-passage flag \\
A5 & TimeBlock & Conversion of abbreviated or relative expressions into comparable temporal markers. & Temporal marker \\
A6 & TimeBlock & Temporal ordering of TimeBlocks on a shared timeline. & Ordered sequence \\
A7 & TimeBlock & Assessment of temporal granularity and suitability as a candidate anchor. & Granularity and anchor candidates \\
A8 & TimeBlock & Cross-text temporal propagation through candidate normalization, episode retrieval, citation verification, within-document anchor stabilization and cross-document constraint construction. & Cross-text evidence and constraints \\
A9 & TimeBlock & Date-range normalization using A8 output and the A6 sequence, including date mapping, range propagation and cross-document soft boundaries. & Final \code{iso} and \code{iso\_range} \\
\bottomrule
\end{tabularx}
\end{table}

\suptabref{tab:supp-nine-agent} summarizes the division of labour. A1--A4 correspond to steps 2--5, A5--A7 to steps 7--9, A8 to steps 10A--10D and A9 to step 11. Within A8, temporal markers are generated and normalized, candidate text pairs are retrieved and checked for event correspondence, citations are verified, within-document anchors are stabilized and cross-document constraints are validated. Step 12 generates readable summaries from the temporal ranges established in step 11.

In the study runs, A8 used cross-document retrieval and large-language-model assessment of event identity, recording candidate relations, cited evidence and verification status. Relations meeting the propagation criteria entered A9, where temporal mapping tables and rule functions generated \code{iso} and \code{iso\_range}; other candidates entered the human-review queue. Cross-text propagation thereby converted verified source evidence into traceable temporal constraints in the target chronology.

\paragraph{Temporal granularity and internal-coordinate serialization.}
A7 assigned temporal information to four levels: \code{0} for an open temporal range, \code{1} for year- or season-level precision, \code{2} for year-and-month precision and \code{3} for year-month-day precision. A8 normalized and verified candidate temporal markers and cross-document evidence. A9 then used reign-title tables, temporal-string mapping tables and rule functions to convert eligible anchors into fixed-width internal date strings. For open temporal ranges, the system searched the A6 sequence for adjacent anchors with non-zero granularity and combined them with cross-document soft boundaries from A8 to construct \code{iso\_range}.

The fields \code{iso} and \code{iso\_range} store AIH internal temporal coordinates. Each endpoint uses a signed, fixed-width \code{YYYY-MM-DD} form. Years and months derive from the Chinese reign-title-to-month mapping used in this study, projecting temporal expressions onto a discrete monthly grid for ordering and MicroIoU calculation. Negative years retain the coordinate values in the mapping table. Outputs preserve the original reign-title expression together with the internal coordinate; a separately curated conversion table supplies calendar-checked dates for standard interchange.

Closed intervals use \code{<start>to<end>}; open endpoints use \code{-infinity} and \code{+infinity}. Displayed results retain the original expression, for example, ``秦二世元年七月'' (seventh month of the first year of Qin Er Shi). The \code{iso\_range} field expresses comparable boundaries formed jointly by the TimeBlock ordering sequence and cross-text evidence.

\subsubsection{Data representation}
\label{sec:supp-data-representation}
\paragraph{Sentence-level fields.}
Sentence-level annotation retains the variables in \suptabref{tab:supp-sentence-fields}, recording people, temporal expressions, textual functions and cross-text evidence relations.

\begin{longtable}{p{0.30\textwidth}p{0.62\textwidth}}
\caption{Principal variables in sentence-level annotation.}\label{tab:supp-sentence-fields}\\
\toprule
Field & Meaning \\
\midrule
\endfirsthead
\toprule
Field & Meaning \\
\midrule
\endhead
\bottomrule
\endfoot
\code{sentence\_id} & Stable sentence identifier in the form \code{book\_uuid.chapter\_id.paragraph\_id.sentence\_id}. \\
\code{text} & Sentence text. \\
\code{character\_flags} & Boolean fields for candidate people such as \code{liu\_bang}, \code{xiang\_yu} and \code{xiao\_he}. \\
\code{has\_time\_expression} & Whether the sentence contains an extractable source temporal expression. \\
\code{time\_expression} & Extracted source wording, such as ``秦二世元年七月'' or ``这时''. \\
\code{is\_sinking} & Whether the sentence is assigned to the background layer of the principal temporal-reasoning chain. \\
\code{sinking\_reason} & Explanation for the sinking decision. \\
\code{is\_interlude} & Whether the sentence belongs to a retrospective passage. \\
\code{crossDocTransfer} & Cross-document propagation record, including cross-text relations and associated TimeBlocks recorded by A8/A9. \\
\code{same\_timeblock\_id} & Identifier of a related TimeBlock in another document. \\
\end{longtable}

\paragraph{TimeBlock fields.}
The TimeBlock representation retains the variables in \suptabref{tab:supp-timeblock-fields}, linking source sentences, temporal markers, normalized ranges and cross-text updates. Field names match the archived JSON output.

\begin{longtable}{p{0.30\textwidth}p{0.62\textwidth}}
\caption{Principal variables in the TimeBlock representation.}\label{tab:supp-timeblock-fields}\\
\toprule
Field & Meaning \\
\midrule
\endfirsthead
\toprule
Field & Meaning \\
\midrule
\endhead
\bottomrule
\endfoot
\code{ID} & Unique TimeBlock identifier, usually derived from the first sentence in the block. \\
\code{timeblock\_range} & Sentence span covered by the TimeBlock. \\
\code{Interlude} & Whether the TimeBlock contains a retrospective passage. \\
\code{Conversion information} & Temporal-marker conversion record, including source wording, conversion requirement, evidence and rationale. \\
\code{Granularity} & Temporal granularity, with values \code{0}--\code{3}. \\
\code{TM} & Normalized temporal marker. \\
\code{iso} & Internal date endpoint normalized by A9; empty at granularity 0. \\
\code{iso\_range} & Internal interval string for the TimeBlock in the ordering sequence. \\
\code{TB\_Update} & Source or description of an update produced during cross-text propagation or soft-boundary processing. \\
\code{summary} & Chinese-language summary generated during output post-processing for timeline cards and retrieval previews. \\
\end{longtable}

\paragraph{Chronological ordering rule.}
The ordering sequence consists of TimeBlock \code{ID}s arranged by temporal logic. When calculating \code{iso\_range}, A9 uses the semantic temporal sequence generated by A6 to identify adjacent anchors and converts cross-document evidence recorded by A8 into soft-boundary or anchor constraints. The sequence is therefore an intermediate representation connecting TimeBlocks to their final temporal ranges.

\section{Evaluation results}
\subsection{End-to-end temporal reconstruction}
\label{sec:supp-experiment1-case-group-results}

\begin{figure}[H]
\centering
\noindent\textbf{a}\par
\includegraphics[width=0.98\textwidth]{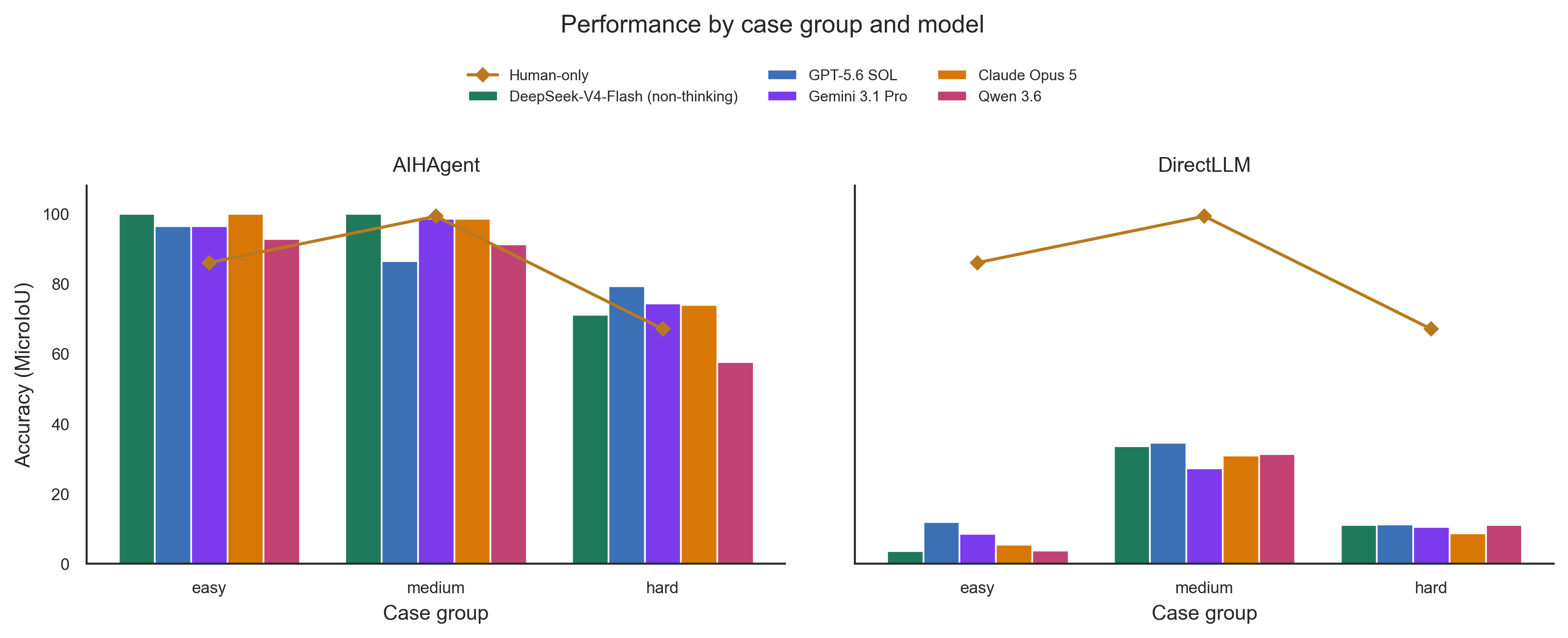}\par\medskip
\noindent\textbf{b}\par
\includegraphics[width=0.98\textwidth]{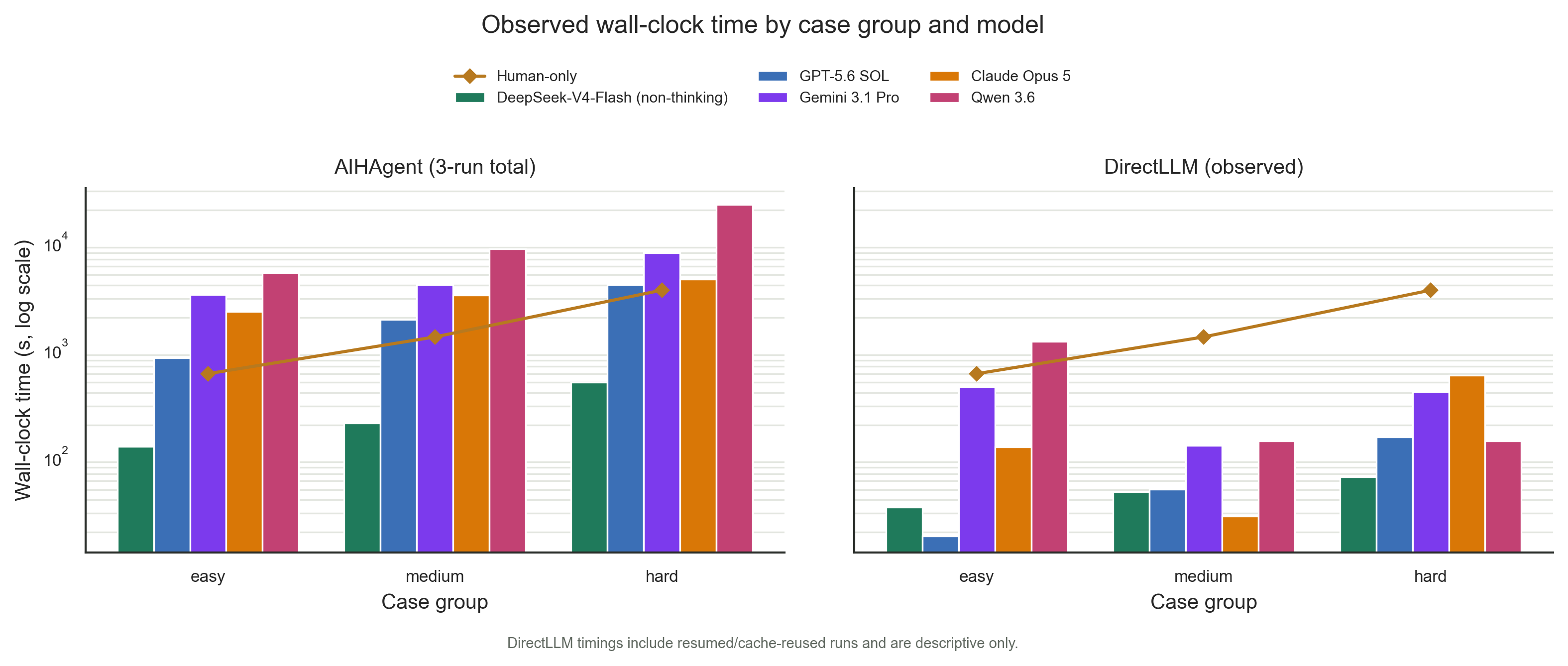}
\caption{Multi-model end-to-end results by case group in Experiment 1. \textbf{a,} Month-level MicroIoU for direct temporal localization (easy; C1--C2), within-document ambiguity reasoning (medium; C3--C4) and cross-document temporal propagation (hard; C5--C6). The two internal panels show AIH Agent and DirectLLM results for DeepSeek-V4-Flash (non-thinking mode), GPT-5.6 Sol, Gemini 3.1 Pro, Claude Opus 5 and Qwen 3.6; the human-only condition is the common reference line. \textbf{b,} Observed wall-clock time for the same groups and models on a logarithmic scale. AIH Agent values sum three independent runs per case; DirectLLM values are the observed times recorded in the run logs.}
\label{fig:supp-experiment1-case-groups}
\end{figure}

\begin{figure}[H]
\centering
\noindent\textbf{a}\par
\includegraphics[width=0.98\textwidth]{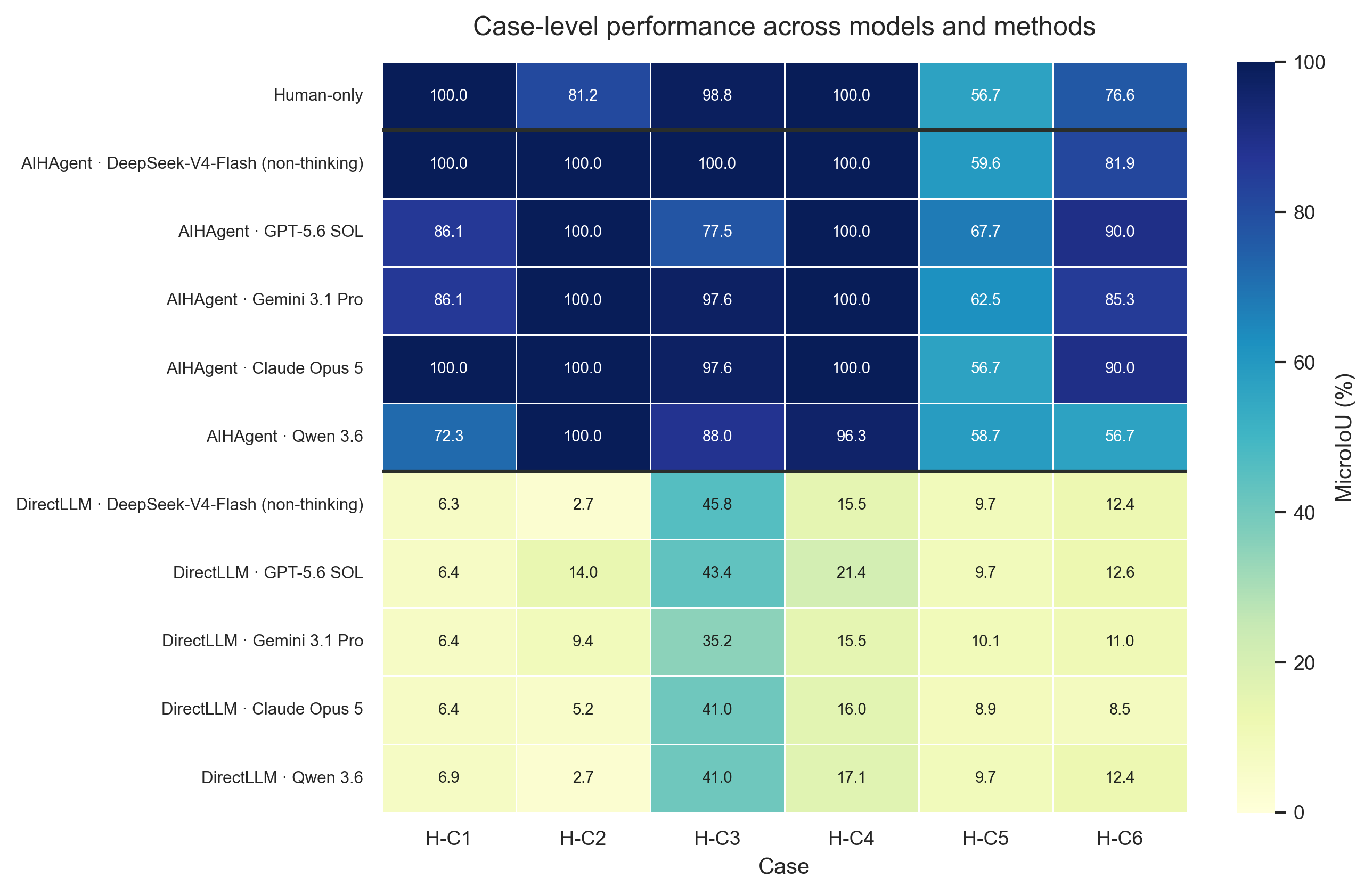}\par\medskip
\noindent\textbf{b}\par
\includegraphics[width=0.98\textwidth]{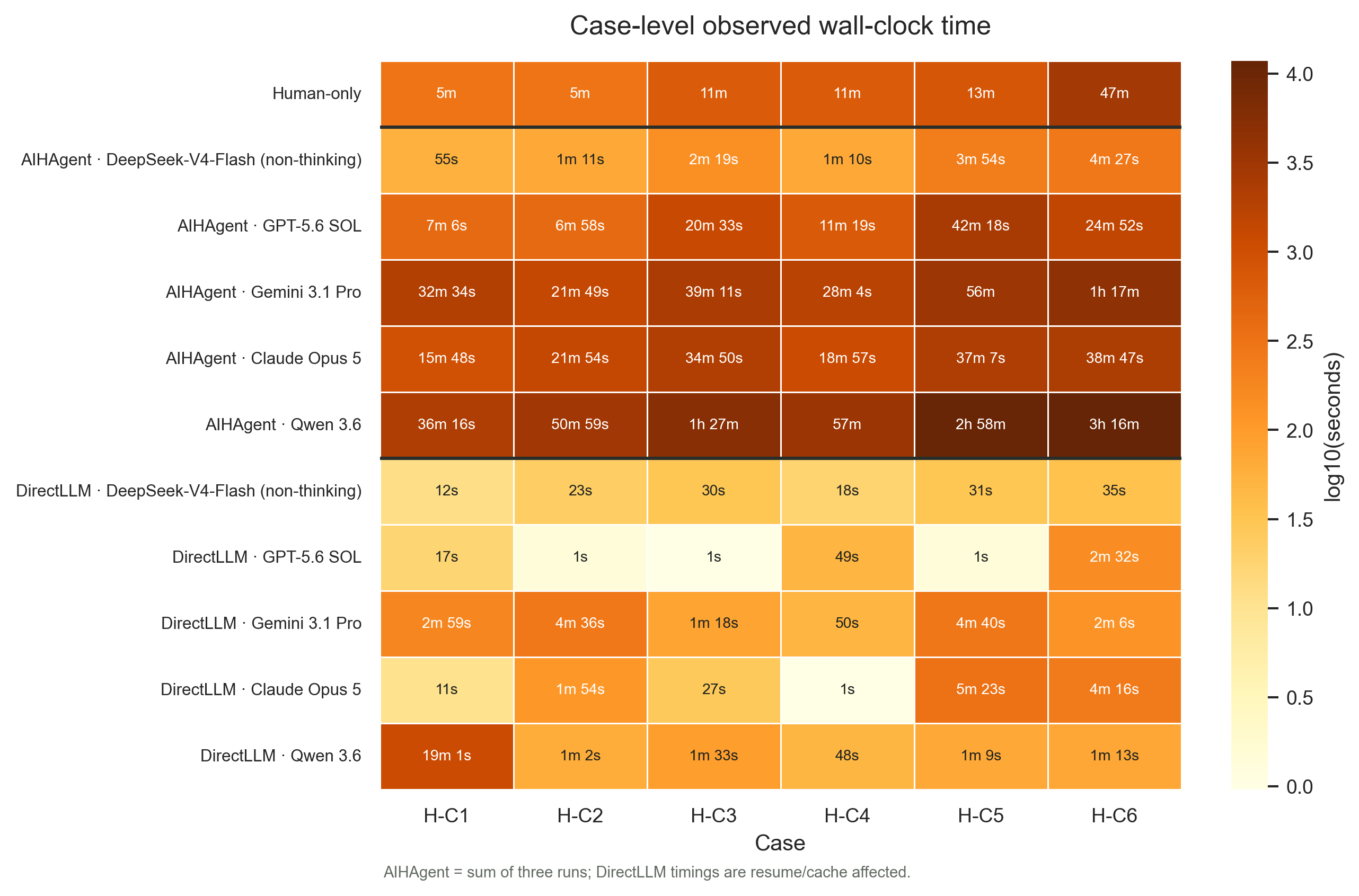}
\caption{Case-level multi-model results in Experiment 1. \textbf{a,} Month-level MicroIoU for the human-only condition and each model--method combination in C1--C6, shown as percentages. \textbf{b,} Observed case-level wall-clock time. AIH Agent cells report the sum of three independent runs; direct-prompting cells report the observed times recorded in the run logs.}
\label{fig:supp-experiment1-case-level}
\end{figure}

The multi-model replication used DeepSeek-V4-Flash in non-thinking mode, GPT-5.6 Sol, Gemini 3.1 Pro, Claude Opus 5 and Qwen 3.6. Across the six cases, descriptive macro-average MicroIoU for AIH Agent was 90.7\% with Claude Opus 5, 90.2\% with DeepSeek-V4-Flash, 88.6\% with Gemini 3.1 Pro, 86.9\% with GPT-5.6 Sol and 78.6\% with Qwen 3.6 (\figref[a]{fig:supp-experiment1-case-level}). Claude Opus 5 exceeded DeepSeek-V4-Flash by 0.5 percentage points. DeepSeek-V4-Flash met or exceeded the corresponding human-only observation in all six cases; GPT-5.6 Sol, Gemini 3.1 Pro and Claude Opus 5 did so in four cases, and Qwen 3.6 in two.

By case group, AIH Agent MicroIoU ranged from 92.9\% to 100.0\% in the easy group, compared with 86.0\% for the human-only observation; from 86.5\% to 100.0\% in the medium group, compared with 99.3\%; and, in the hard group, reached 79.4\% for GPT-5.6 Sol, 74.4\% for Gemini 3.1 Pro, 74.1\% for Claude Opus 5, 71.2\% for DeepSeek-V4-Flash and 57.6\% for Qwen 3.6, compared with 67.1\% (\figref[a]{fig:supp-experiment1-case-groups}). Direct prompting yielded 3.7\%--12.0\%, 27.3\%--34.6\% and 8.7\%--11.2\% in the easy, medium and hard groups, respectively. At case level, GPT-5.6 Sol was highest in C5 (67.7\%); GPT-5.6 Sol and Claude Opus 5 tied in C6 (90.0\%); and DeepSeek-V4-Flash reached 100.0\% in C3.

Observed times are shown in \figref[b]{fig:supp-experiment1-case-groups} and \figref[b]{fig:supp-experiment1-case-level}. Human-only values derive from manual timing, AIH Agent values sum three independent runs per case and direct-prompting values derive from the corresponding run logs.

\subsection{Component-level diagnostics}
\label{sec:supp-diagnostic-results}

\begin{figure}[H]
\centering
\begin{minipage}[t]{0.32\textwidth}
\textbf{a}\par\centering
\includegraphics[width=\linewidth]{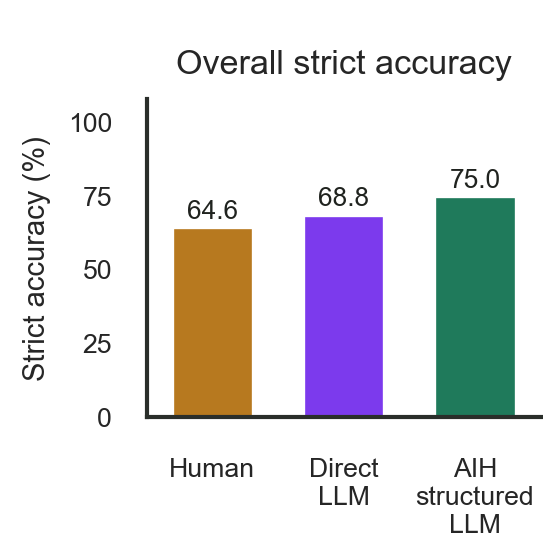}
\end{minipage}\hfill
\begin{minipage}[t]{0.64\textwidth}
\textbf{b}\par\centering
\includegraphics[width=\linewidth]{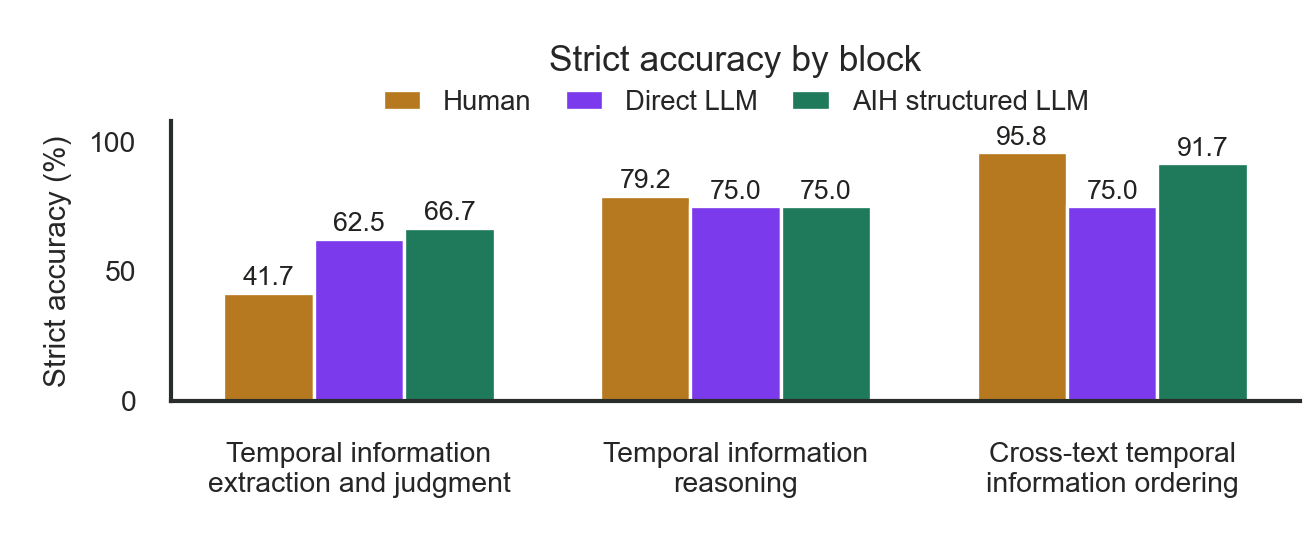}
\end{minipage}
\caption{Diagnostic accuracy in Experiment 2. \textbf{a,} Overall strict accuracy for human responses, direct prompting and structured prompting across 48 unique diagnostic questions. \textbf{b,} Strict accuracy for temporal-information extraction and judgement, temporal-information reasoning, and cross-text event verification and temporal alignment. The figure label ``cross-text temporal-information ordering'' encompasses cross-text candidate-relation and temporal-alignment judgements. Human values aggregate 96 responses; model values are aggregated over three runs of the same 48 questions.}
\label{fig:supp-diagnostic-accuracy}
\end{figure}

\begin{figure}[H]
\centering
\begin{minipage}[t]{0.32\textwidth}
\textbf{a}\par\centering
\includegraphics[width=\linewidth]{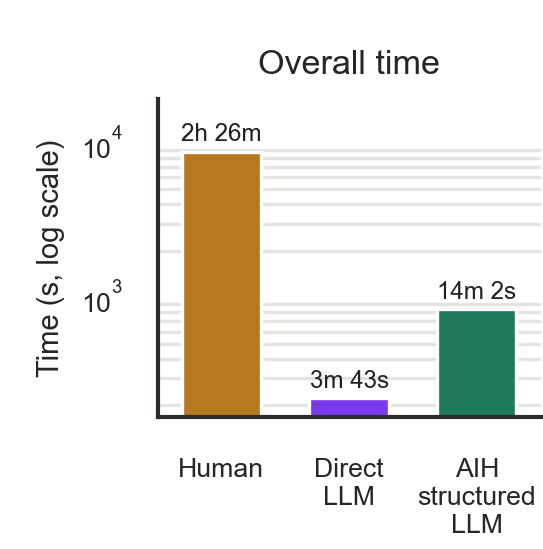}
\end{minipage}\hfill
\begin{minipage}[t]{0.64\textwidth}
\textbf{b}\par\centering
\includegraphics[width=\linewidth]{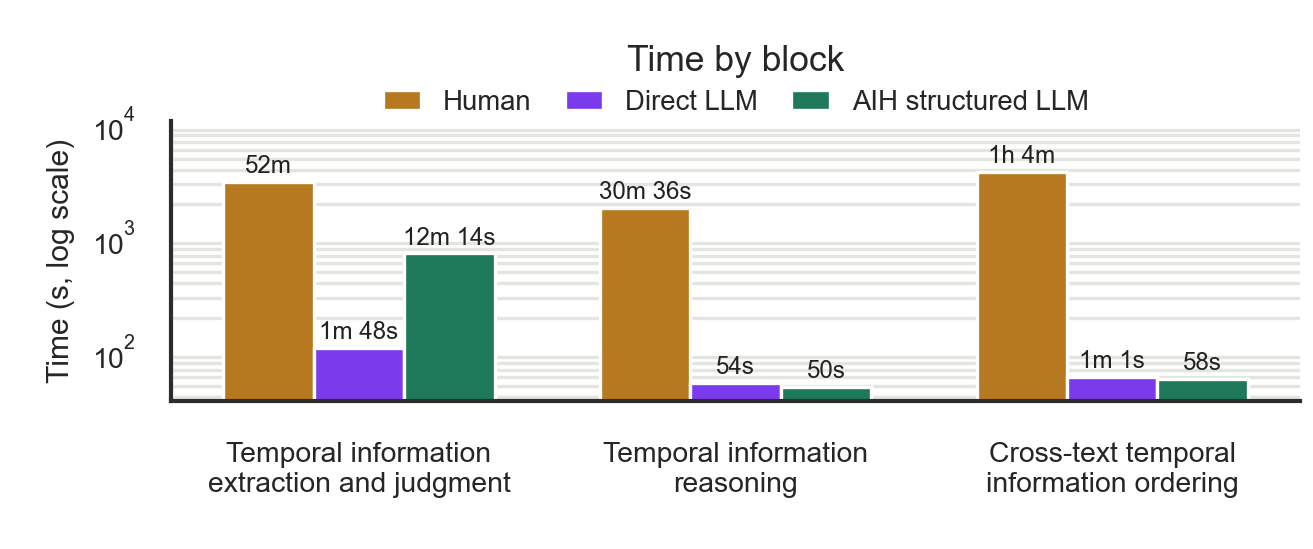}
\end{minipage}
\caption{Diagnostic elapsed time in Experiment 2. \textbf{a,} Cumulative elapsed time for human responses, direct prompting and structured prompting. \textbf{b,} Elapsed time for temporal-information extraction and judgement, temporal-information reasoning, and cross-text event verification and temporal alignment. The figure label ``cross-text temporal-information ordering'' encompasses cross-text candidate-relation and temporal-alignment judgements. Time is shown in seconds on a logarithmic scale.}
\label{fig:supp-diagnostic-time}
\end{figure}

The diagnostic experiment examined the contribution of structured intermediate representations to component-level judgements by comparing human responses, direct LLM prompts and diagnostic prompts derived from AIH intermediate steps. Overall strict accuracy was 75.0\% (36/48) for structured prompting, 68.8\% (33/48) for direct prompting and 64.6\% (62/96) for human responses (\figref[a]{fig:supp-diagnostic-accuracy}). Cumulative elapsed time was 14\,min 2\,s, 3\,min 43\,s and 2\,h 26\,min, respectively. Direct prompting had the shortest recorded time, while structured prompting had the highest strict accuracy (\figref[a]{fig:supp-diagnostic-time}).

The main difference between structured and direct prompting appeared in the cross-text diagnostic module (\figref[b]{fig:supp-diagnostic-accuracy}). For temporal-information extraction and judgement, including recognition of temporal expressions, background statements and retrospective passages, strict accuracy was 66.7\%, 62.5\% and 41.7\% for structured prompting, direct prompting and human responses. For temporal-information reasoning, including temporal-marker completion and TimeBlock ordering, structured and direct prompting both reached 75.0\%, and human responses reached 79.2\%. For cross-text event verification and temporal alignment, the corresponding values were 91.7\%, 75.0\% and 95.8\%.

Module-level elapsed times are shown in \figref[b]{fig:supp-diagnostic-time}. For temporal-information extraction and judgement, cumulative times were 52\,min, 1\,min 48\,s and 12\,min 14\,s for human responses, direct prompting and structured prompting. For temporal-information reasoning, they were 30\,min 36\,s, 54\,s and 50\,s. For cross-text event verification and temporal alignment, they were 1\,h 4\,min, 1\,min 1\,s and 58\,s. The largest observed accuracy gain of structured over direct prompting occurred in cross-text candidate-relation judgement and temporal alignment.

\section{Cross-text cases}
\subsection{Evidence propagation in C5--C6}
\label{sec:supp-cross-textual-propagation-case}

C5 and C6 evaluate cross-text event verification and temporal alignment. A passage with an explicit temporal anchor in one biography constrains a phase-level account in another, and the system generates final person-level temporal ranges from candidate text relations and temporal boundaries.

\paragraph{Evaluation passages and annotation scope.}
The cross-text annotation covers two sets of material: C5 links the \textit{Basic Annals of Xiang Yu} and the \textit{Basic Annals of Gaozu}, while C6 links the \textit{Basic Annals of Gaozu} and the \textit{Hereditary House of Chancellor Xiao}. Sentences describing the same historical process or adjacent historical phases across biographies are placed on a shared, comparable timeline.

\begin{table}[H]
\centering
\caption{Examples of internal reference temporal coordinates in the cross-text annotation.}
\label{tab:supp-crossdoc-reference-examples}
\small
\begin{tabularx}{\textwidth}{@{}p{0.10\textwidth}p{0.13\textwidth}p{0.15\textwidth}p{0.34\textwidth}p{0.18\textwidth}@{}}
\toprule
Case & Sentence & Chapter & Passage & Internal reference range \\
\midrule
C5 & \code{8.16.1} & Gaozu & In the third year of Qin Er Shi, King Huai of Chu saw that Xiang Liang's army had been defeated, moved his capital to Pengcheng and combined the armies of Lü Chen and Xiang Yu. & \code{-0207-10to-0206-11} \\
C5 & \code{7.10.1} & Xiang Yu & After the Chu army was defeated at Dingtao, King Huai moved to Pengcheng, combined the armies of Xiang Yu and Lü Chen and assumed command. & \code{-0207-10to-0206-11} \\
C5 & \code{8.23.3} & Gaozu & In the eleventh month of the first Han year, Xiang Yu led the armies of the feudal lords westward to enter the passes, but the gate was closed. & \code{-0206-11to-0206-12} \\
C5 & \code{7.17.3} & Xiang Yu & Xiang Yu heard that Pei Gong had taken Xianyang, became enraged and sent Lord Dangyang and others to attack the pass. & \code{-0206-11to-0206-12} \\
C6 & \code{53.4.1} & Xiao He & After Liu Bang rose and became Pei Gong, Xiao He frequently assisted him in administering affairs. & \code{-0209-10to-0208-10} \\
C6 & \code{53.4.2} & Xiao He & When Pei Gong entered Xianyang, Xiao He first secured the Qin laws, maps and household registers. & \code{-0208-10to-0205-01} \\
C6 & \code{53.4.3} & Xiao He & When Pei Gong became King of Han, he appointed Xiao He as chancellor. & \code{-0205-01to-0205-10} \\
C6 & \code{8.26.2} & Gaozu & Xiang Yu broke the earlier agreement and made Pei Gong King of Han, ruling Ba, Shu and Hanzhong from Nanzheng. & \code{-0205-01to-0205-04} \\
\bottomrule
\end{tabularx}
\end{table}

\paragraph{A8 cross-text evidence verification.}
In C5, the \textit{Basic Annals of Gaozu} and the \textit{Basic Annals of Xiang Yu} record, from the perspectives of Liu Bang and Xiang Yu, the same sequence in which King Huai of Chu reorganized his forces, assigned westward campaigns and Xiang Yu encountered the closed pass. Sentences \code{8.16.1} and \code{7.10.1}, for example, both describe King Huai's move to Pengcheng and consolidation of the armies after Xiang Liang's defeat; the reference annotation assigns both to \code{-0207-10to-0206-11}. Likewise, \code{8.23.3} and \code{7.17.3} record the phase in which Xiang Yu's westward advance was blocked and he reacted to Pei Gong's capture of the region within the passes; both receive \code{-0206-11to-0206-12}.

In C6, the \textit{Hereditary House of Chancellor Xiao} summarizes Xiao He's experience from Liu Bang's uprising through the entry into Xianyang, his appointment as chancellor and his administration of Ba and Shu. The annotation places these statements on the Chu--Han timeline supplied by the \textit{Basic Annals of Gaozu}: \code{53.4.1} receives \code{-0209-10to-0208-10}, \code{53.4.2} receives \code{-0208-10to-0205-01} and \code{53.4.3} receives \code{-0205-01to-0205-10}. A8 verifies the event-context relation between the summary statements in Xiao He's biography and the more specific political and military phases in Gaozu's biography.

\paragraph{A9 temporal-range normalization and propagation.}
A9 converts the normalized source evidence recorded by A8 into temporal constraints in the target text. C5 illustrates evidence propagation within a shared phase: verified candidate pairs describing King Huai's military reorganization and Xiang Yu's passage through the gates receive the same or closely corresponding reference ranges. C6 illustrates biographical completion: sentences in the \textit{Hereditary House of Chancellor Xiao} correspond to the uprising, entry into Xianyang, enfeoffment as King of Han and eastward campaign in the \textit{Basic Annals of Gaozu}. A9 generates \code{iso\_range} by combining the target-sentence order, adjacent boundaries and cross-text phase constraints.

\begin{table}[H]
\centering
\caption{Cross-text internal reference coordinates for Xiao He passages in C6.}
\label{tab:supp-xiaohe-timeblock}
\small
\begin{tabularx}{\textwidth}{@{}p{0.16\textwidth}p{0.46\textwidth}p{0.25\textwidth}@{}}
\toprule
Sentence & Passage & Internal reference range \\
\midrule
\code{53.4.1} & After Liu Bang rose and became Pei Gong, Xiao He repeatedly assisted him in administering affairs. & \code{-0209-10to-0208-10} \\
\code{53.4.2} & When Pei Gong entered Xianyang, Xiao He first secured the laws, maps and household registers. & \code{-0208-10to-0205-01} \\
\code{53.4.3} & When Pei Gong became King of Han, he appointed Xiao He as chancellor. & \code{-0205-01to-0205-10} \\
\code{53.5.1} & When the King of Han led his army east and pacified the Three Qin, Xiao He remained as chancellor to govern Ba and Shu. & \code{-0205-10to+infinity} \\
\bottomrule
\end{tabularx}
\end{table}

Cross-text propagation combines A8 evidence verification with A9 temporal-constraint generation. A8 establishes event relations between source and target TimeBlocks and records citations, anchors and interval boundaries. A9 updates the target chronology according to relation type: same-event relations provide transferable anchors, containment relations provide source intervals, and shared historical phases provide preceding and following boundary constraints. For target TimeBlocks with open ranges, these constraints enter the \code{iso\_range} calculation and update the matched phase and adjacent intervals. C5 and C6 test whether a shared historical phase distributed across biographies enters a consistent temporal structure.

\section{Platform and structured outputs}
\subsection{Platform workflow}
\label{sec:supp-platform-functions}

Westlake Historian implements the historian-led workflow summarized in Fig.~4. Users can upload biographical texts and inspect monitored AIH processing, retrieve evidence by person and temporal scope, compare source passages, ask source-grounded questions, translate chronological materials, generate readable biographical drafts and trace a time point to its source sentence and temporal basis. These functions share the AIH structured-evidence layer.

The platform also records post-generation review and publication. A chronology is first stored in its owner's private workspace. Users may submit revisions with reasons, receive an AI-assisted review recommendation and retain the final human decision. Accepted versions may remain private or be published to the Chronology Market for discovery, inspection and discussion. The following interfaces document the implemented workflow.

\begin{figure}[H]
\centering
\includegraphics[width=0.95\textwidth]{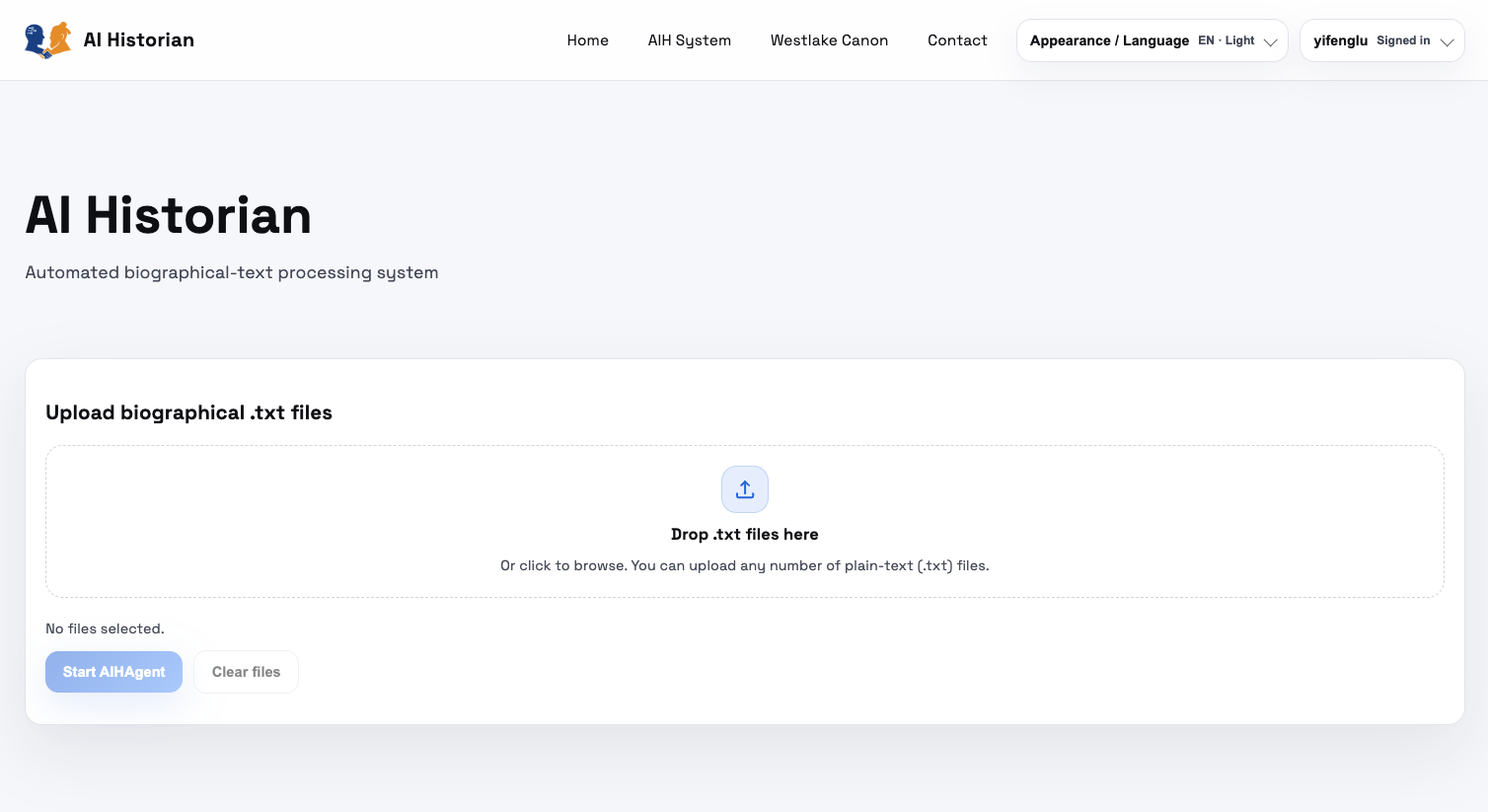}
\caption{Text upload and monitored AIH processing. Users upload biographical texts and inspect the process that generates source-traceable chronology and temporal-range views.}
\label{fig:supp-aih-interface}
\label{fig:supp-platform-upload}
\end{figure}

\begin{figure}[H]
\centering
\includegraphics[width=0.95\textwidth]{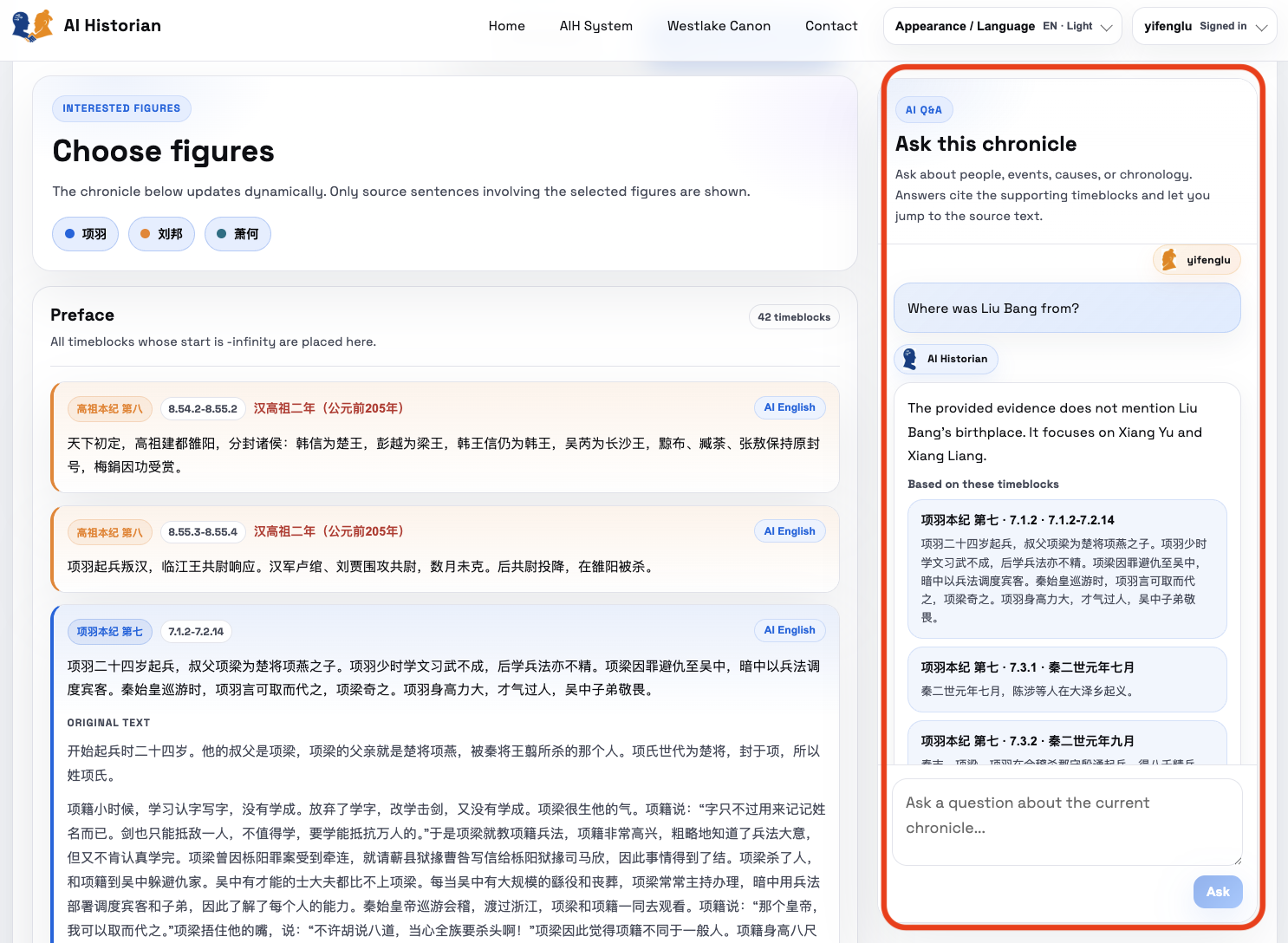}
\caption{Source-grounded question answering. The interface supports questions over AIH-structured historical material while retaining access to the underlying evidence.}
\label{fig:supp-platform-qa}
\end{figure}

\begin{figure}[H]
\centering
\includegraphics[width=0.95\textwidth]{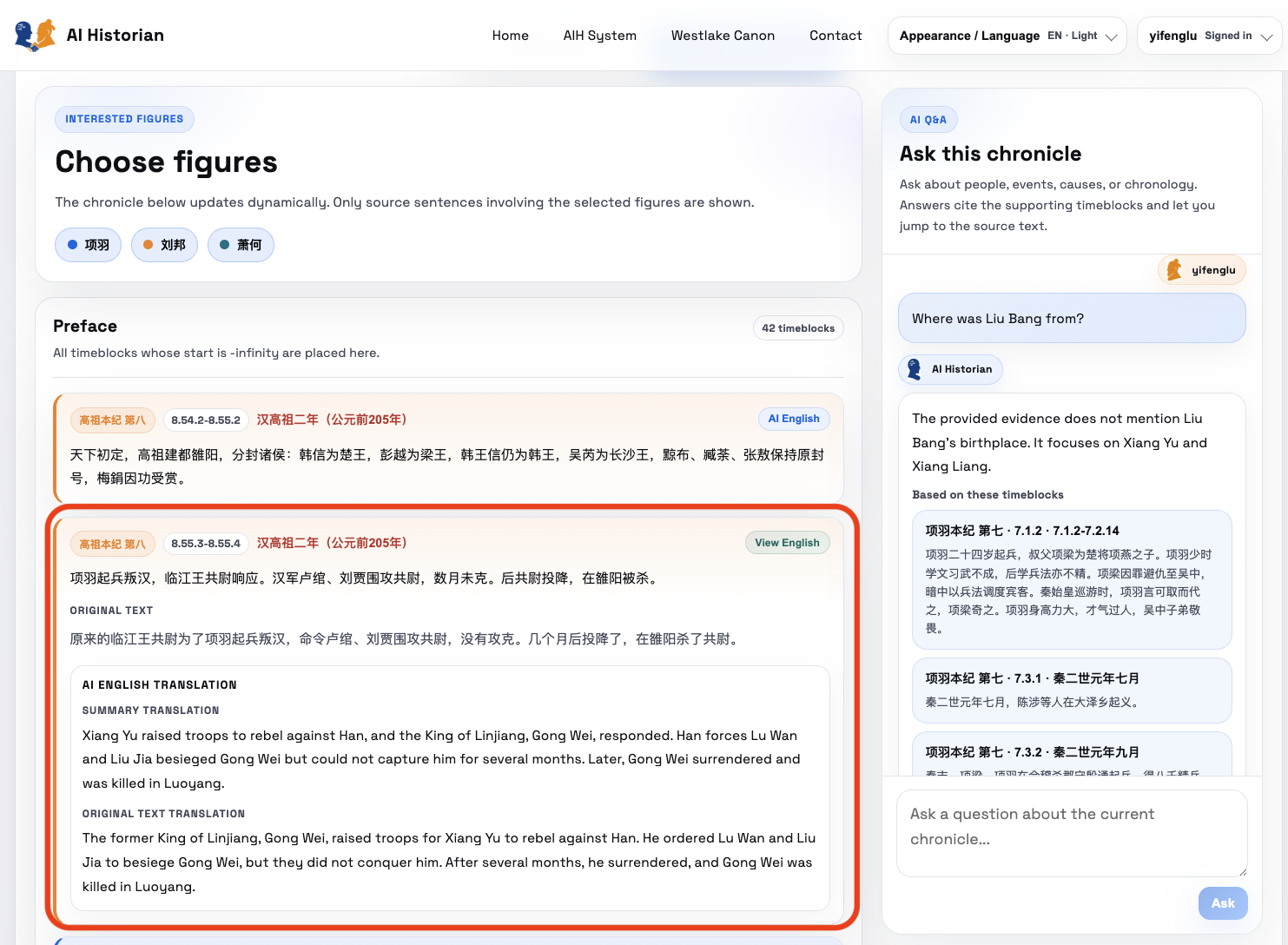}
\caption{Bilingual translation. The interface provides AI-assisted Chinese--English translation to support reading of AIH-organized chronological material.}
\label{fig:supp-platform-translation}
\end{figure}

\begin{figure}[H]
\centering
\includegraphics[width=0.95\textwidth]{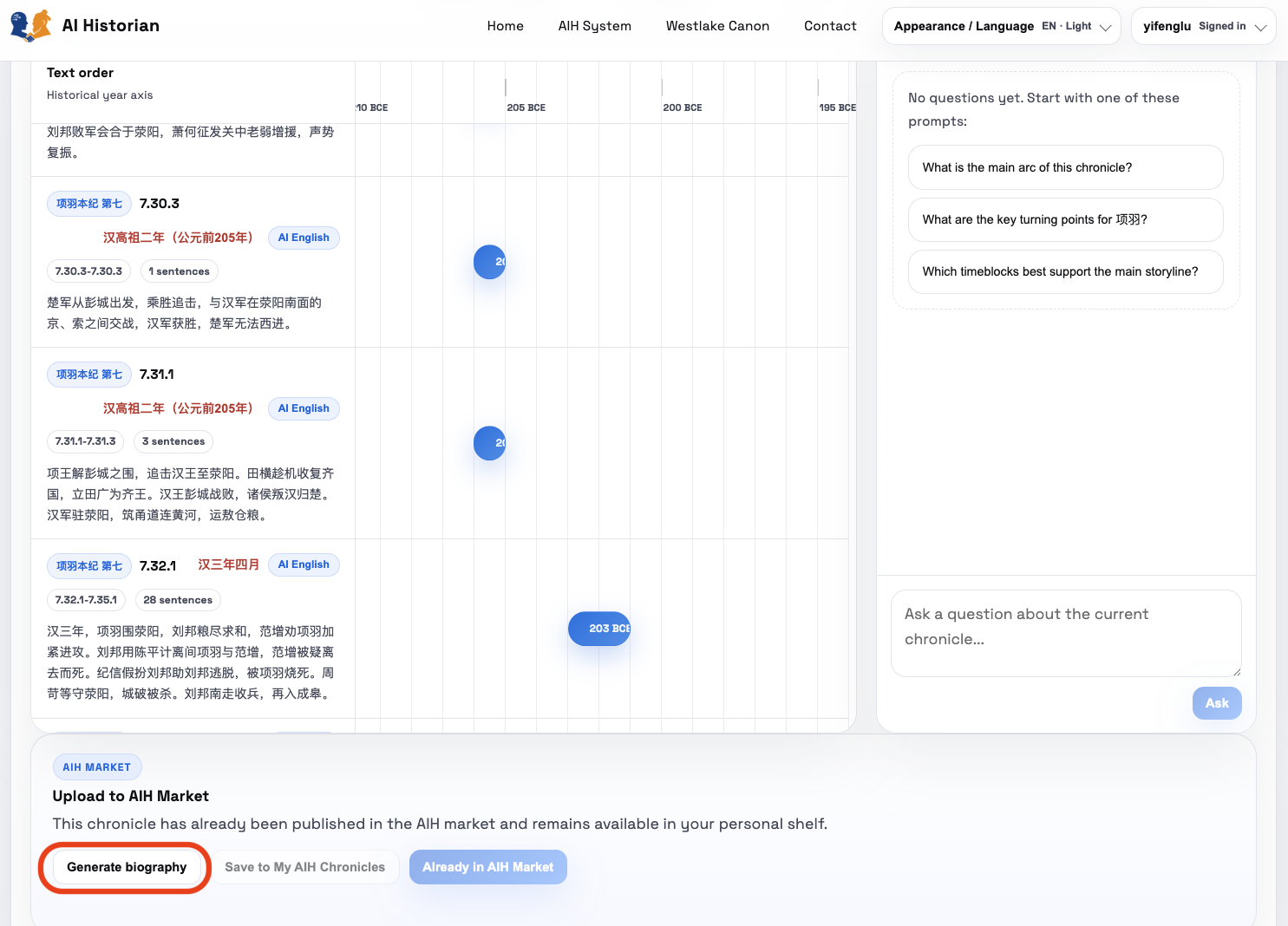}
\caption{Biographical-draft generation. The interface generates a readable draft from AIH-structured chronological material for continued scholarly review and revision.}
\label{fig:supp-platform-biography}
\end{figure}

\begin{figure}[H]
\centering
\includegraphics[width=0.95\textwidth]{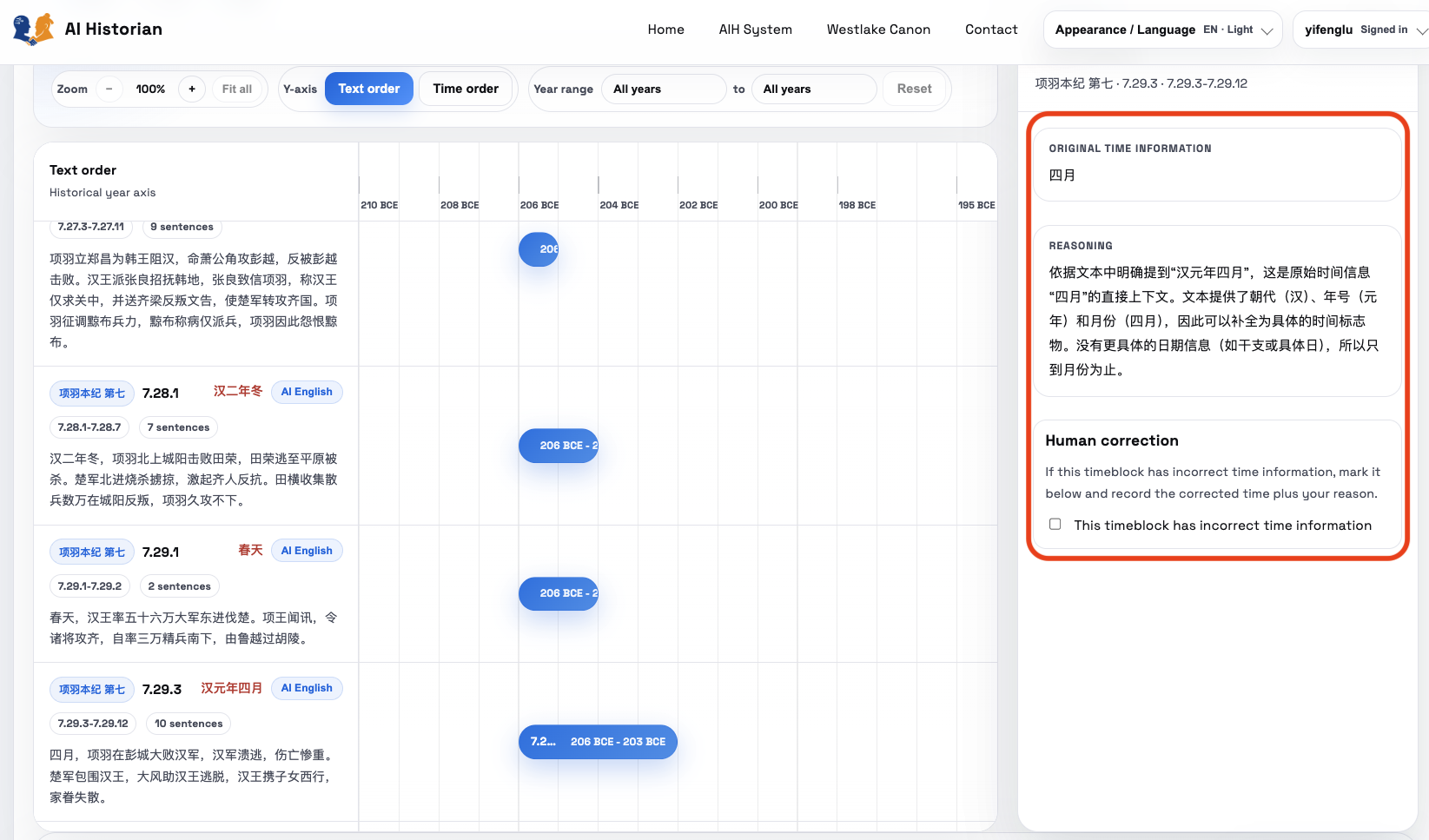}
\caption{Time-point source inspection. A chronological point can be traced to its source sentence and temporal evidence for examination by historians.}
\label{fig:supp-platform-time-source}
\end{figure}

\begin{figure}[H]
\centering
\includegraphics[width=0.95\textwidth]{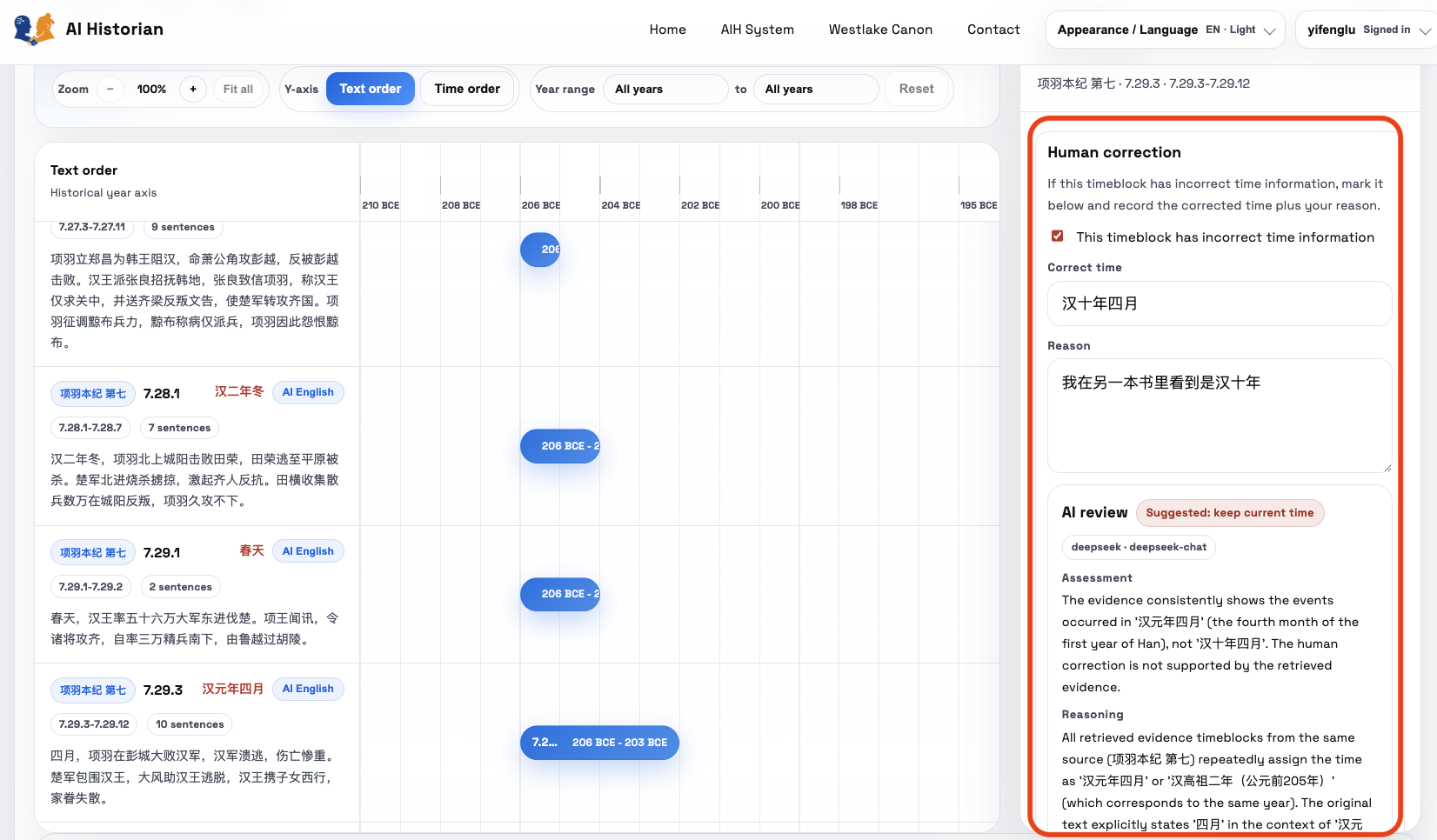}
\caption{AI-assisted revision review. The interface examines a submitted revision and its rationale, returns an AI recommendation and records the final human judgement.}
\label{fig:supp-platform-review}
\end{figure}

\begin{figure}[H]
\centering
\includegraphics[width=0.95\textwidth]{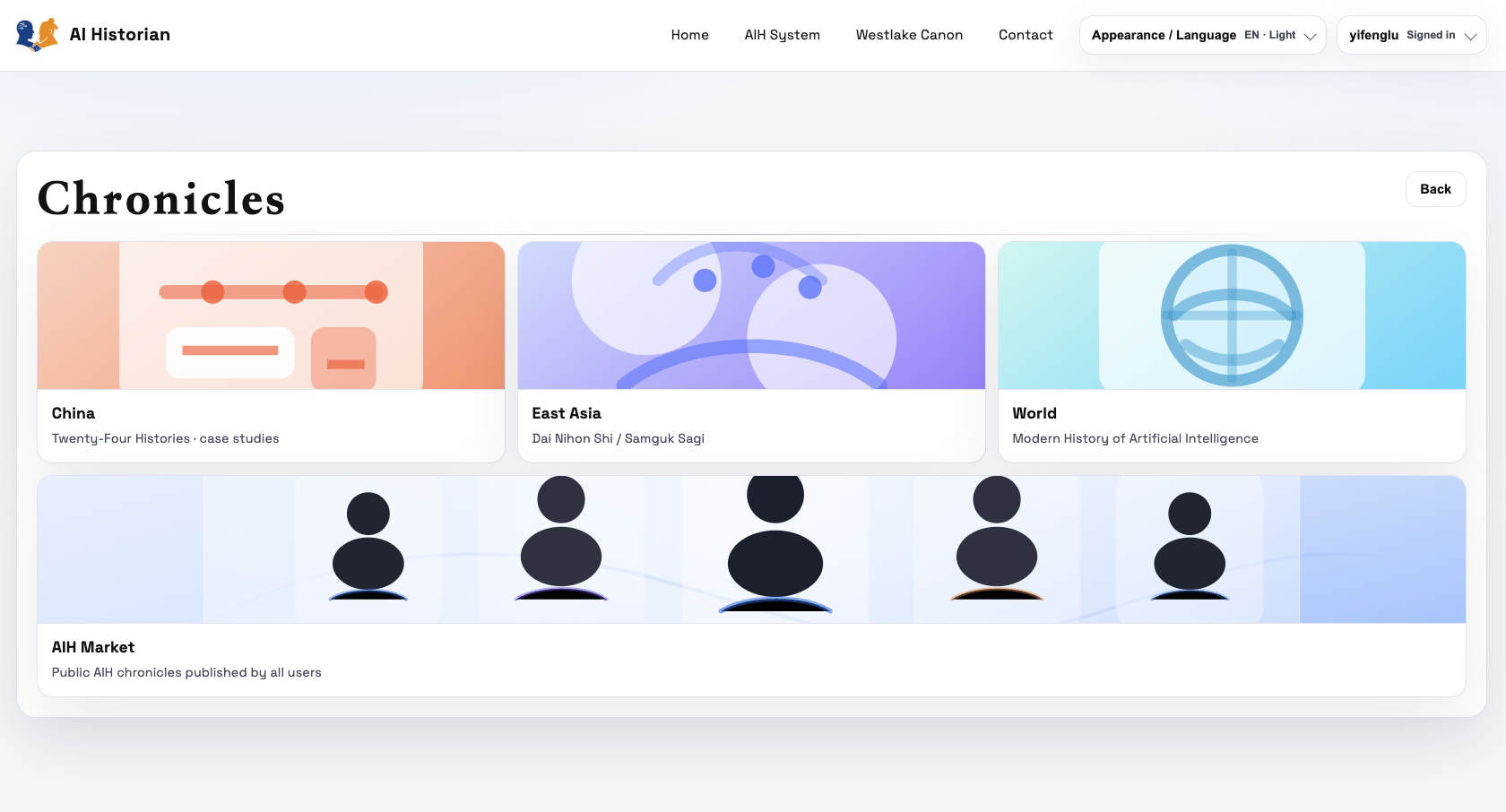}
\caption{Platform corpus coverage. The interface lists processed corpus collections and document tracks spanning East Asian historiographical materials and modern histories of science and technology.}
\label{fig:supp-platform-coverage}
\end{figure}

\subsection{Corpus coverage and outputs}
\suptabref{tab:supp-platform-coverage} summarizes the materials processed by Westlake Historian and their structured outputs as of 10 July 2026. Cross-text relation records comprise 91 strong relations (\textit{episode\_context} or \textit{same\_sequence\_phase}) and 684 candidate relations. Each record preserves passages from the source and target texts; strong relations enter A9 temporal-constraint propagation, and candidate relations enter subsequent human review.

{\scriptsize
\setlength{\tabcolsep}{2pt}
\begin{longtable}{P{0.27\textwidth}rrrrrr}
\caption{Westlake Historian corpus coverage and structured outputs as of 10 July 2026.}\label{tab:supp-platform-coverage}\\
\toprule
Corpus collection & Texts & Sentences & TimeBlocks & \shortstack{Temporal\\anchors} & \shortstack{Normalized\\ranges} & \shortstack{Cross-text\\relations} \\
\midrule
\endfirsthead
\toprule
Corpus collection & Texts & Sentences & TimeBlocks & \shortstack{Temporal\\anchors} & \shortstack{Normalized\\ranges} & \shortstack{Cross-text\\relations} \\
\midrule
\endhead
\bottomrule
\endfoot
\textit{Samguk Sagi} & 3 & 73 & 23 & 20 & 23 & 19 \\
\textit{Sanguozhi} & 3 & 345 & 57 & 19 & 57 & 10 \\
\textit{History of Yuan} & 3 & 7,187 & 3,405 & 950 & 3,405 & 36 \\
\textit{History of the Northern Dynasties} & 3 & 930 & 174 & 66 & 174 & 26 \\
\textit{Book of Northern Qi} & 3 & 639 & 216 & 122 & 216 & 57 \\
\textit{History of the Southern Dynasties} & 3 & 433 & 30 & 1 & 30 & 14 \\
\textit{Book of Southern Qi} & 3 & 853 & 459 & 221 & 459 & 35 \\
\textit{Shiji} & 3 & 1,100 & 114 & 83 & 114 & 35 \\
\textit{Book of the Later Han} & 3 & 1,217 & 392 & 212 & 392 & 25 \\
\textit{Book of Zhou} & 3 & 1,560 & 709 & 309 & 709 & 38 \\
\textit{Dai Nihonshi} & 3 & 7,208 & 307 & 20 & 307 & 62 \\
\textit{Book of Song} & 3 & 1,459 & 632 & 207 & 632 & 17 \\
\textit{History of Song} & 3 & 2,428 & 1,852 & 683 & 1,852 & 45 \\
Old and New Histories of the Five Dynasties & 3 & 3,253 & 1,627 & 541 & 1,627 & 44 \\
Old and New Books of Tang & 3 & 1,936 & 1,004 & 445 & 1,004 & 26 \\
\textit{History of Ming} & 3 & 2,258 & 1,578 & 787 & 1,578 & 25 \\
\textit{Book of Jin} & 3 & 1,366 & 696 & 330 & 696 & 12 \\
\textit{Book of Liang} & 3 & 1,281 & 856 & 448 & 856 & 28 \\
\textit{Book of Han} & 3 & 680 & 268 & 164 & 268 & 25 \\
\textit{History of Liao} & 3 & 1,402 & 924 & 417 & 924 & 33 \\
Modern history of AI (Turing, Post and Newman) & 3 & 53 & 32 & 23 & 32 & 6 \\
\textit{History of Jin} & 3 & 1,540 & 889 & 395 & 889 & 45 \\
\textit{Book of Chen} & 3 & 972 & 649 & 286 & 649 & 40 \\
\textit{Book of Sui} & 3 & 1,800 & 958 & 441 & 958 & 35 \\
\textit{Book of Wei} & 3 & 2,339 & 1,566 & 712 & 1,566 & 37 \\
\midrule
Total & 75 & 44,312 & 19,417 & 7,902 & 19,417 & 775 \\
\end{longtable}
}

The 25 corpus collections cover 26 historical works and one collection on the modern history of artificial intelligence. The two comparative collections combining the old and new histories each contain two historical works, so the number of historical works exceeds the number of corpus collections. Each text corresponds to one person timeline in the platform, yielding the 75 person timelines reported in the main text.

The API cost range in the main text was calculated from the recorded call volume using the public list prices for Qwen3.6-35B-A3B (RMB 1.8 and RMB 10.8 per million input and output tokens, respectively). Pricing is documented at \url{https://help.aliyun.com/zh/model-studio/qwen3-6-35b-a3b} (accessed 18 August 2026).

\subsection{Structured-output examples}
\label{sec:supp-structured-output}
\paragraph{Output statistics for the three \textit{Shiji} chapters.}
The final AIH output is a set of structured TimeBlocks. Each TimeBlock preserves the source-sentence span, temporal marker, normalized temporal range generated by A9, cross-text update record and summary field, supporting timeline display, evidence tracing and human revision. A9 determines the temporal range in step 11, and step 12 generates the display-layer summary. \suptabref{tab:supp-platform-coverage} reports the platform snapshot of 10 July 2026; the following table summarizes archived whole-chapter outputs for the three \textit{Shiji} chapters. Experiment 1 scores the 244 sentence-level temporal ranges in the six case packets.

\begin{table}[H]
\centering
\caption{Final TimeBlock output statistics for the three \textit{Shiji} chapters.}
\label{tab:supp-timeblock-overview}
\small
\begin{tabularx}{\textwidth}{Xrrrr}
\toprule
Chapter & TimeBlocks & Non-empty ISO & Cross-text update records & Summaries \\
\midrule
\textit{Basic Annals of Xiang Yu} & 88 & 77 & 57 & 88 \\
\textit{Basic Annals of Gaozu} & 166 & 139 & 85 & 166 \\
\textit{Hereditary House of Chancellor Xiao} & 30 & 23 & 12 & 30 \\
\midrule
Total & 284 & 239 & 154 & 284 \\
\bottomrule
\end{tabularx}
\end{table}

\suptabref{tab:supp-timeblock-overview} shows the structured output after processing the three full chapters. The TimeBlocks can be queried, ordered, filtered and traced to source evidence. A total of 154 TimeBlocks contain cross-text update records preserving candidate relations, evidence or propagation status. Relations that pass evidence verification and meet propagation criteria are converted by A9 into final temporal constraints.

\paragraph{Representative TimeBlock entries.}
\suptabref{tab:supp-final-excerpts} presents representative TimeBlocks, retaining their \code{ID}, normalized time and \code{summary} fields.

\begin{longtable}{@{}p{0.18\textwidth}p{0.18\textwidth}p{0.18\textwidth}p{0.38\textwidth}@{}}
\caption{Representative TimeBlock output entries.}\label{tab:supp-final-excerpts}\\
\toprule
Chapter & \code{ID} & Normalized time & \code{summary} \\
\midrule
\endfirsthead
\toprule
Chapter & \code{ID} & Normalized time & \code{summary} \\
\midrule
\endhead
\bottomrule
\endfoot
Xiang Yu & \code{7.3.1} & \code{-0209-07-01} & In the seventh month of the first year of Qin Er Shi, Chen She and others launched the uprising at Daze Village, beginning the late-Qin rebellions. \\
Xiang Yu & \code{7.5.14} & \code{-0209-09-01} & Pei Gong raised troops at Pei and led an attack on Xue. \\
Gaozu & \code{8.23.3} & \code{-0206-11to-0206-12} & In the eleventh month of the first Han year, Xiang Yu led the armies of the feudal lords west and was blocked at Hangu Pass. \\
Gaozu & \code{8.26.2} & \code{-0205-01to-0205-04} & Xiang Yu made Pei Gong King of Han, ruling Ba, Shu and Hanzhong from Nanzheng. \\
Xiao He & \code{53.4.1} & \code{-0209-10to-0208-10} & After Liu Bang rose as Pei Gong, Xiao He repeatedly assisted him in administration and became a principal aide. \\
Xiao He & \code{53.4.2} & \code{-0208-10to-0205-01} & When Pei Gong entered Xianyang, Xiao He first preserved the laws, maps, registers and archival materials. \\
\end{longtable}

In the complete JSON, these entries retain \code{ID}, \code{timeblock\_range}, \code{TM}, \code{iso}, \code{iso\_range} and \code{TB\_Update}. Each final output can serve as a timeline unit and can be traced through the derived annotations that produced it.

\section{Terminology and proper names}
\subsection{Method terminology}
Concepts denoting specific historiographical forms or AIH components retain their Chinese source terms. After first definition, the English manuscript uses the corresponding English or romanized form. \suptabref{tab:supp-terminology} lists the adopted terminology.

\begin{longtable}{P{0.20\textwidth}P{0.29\textwidth}P{0.43\textwidth}}
\caption{Chinese and English terminology used in the manuscript.}\label{tab:supp-terminology}\\
\toprule
\textbf{Chinese term} & \textbf{English usage} & \textbf{Definition in this study} \\
\midrule
\endfirsthead
\toprule
\textbf{Chinese term} & \textbf{English usage} & \textbf{Definition in this study} \\
\midrule
\endhead
\bottomrule
\endfoot
人物编年（年谱） & biographical chronology & Chronological organization of a person's life, activities and relationships; \textit{nianpu} is the traditional Chinese form discussed here. \\
传统人物年谱 & traditional biographical chronology & Concise chronology centred on the principal events of a person's life. \\
年谱长编 & extended biographical chronology & Evidence-rich chronology incorporating source excerpts, quotations and editorial notes. \\
多人物合谱 & multi-person chronology & Combined chronology representing interactions among several historical people. \\
编年 & chronology & Events arranged by time; here, a method or temporal structure. \\
编年体 & annalistic format & Historiographical format organized primarily by chronological sequence. \\
纪传体 & annals-biographies format & Historiographical format organized through basic annals, biographies and thematic chapters, as in the \textit{Shiji}. \\
时间块 & TimeBlock & Contiguous text governed by a common temporal marker and used as the basic temporal-reasoning unit. \\
时间标志物 & temporal marker & Textual expression that anchors or constrains the time of a TimeBlock. \\
时间标志物缓存区 & Temporal Marker Buffer (TMB) & Document-level memory structure that normalizes elliptical or relative temporal expressions. \\
时间信息原文 & source temporal expression & Original wording of a temporal expression before normalization. \\
下沉文本 & sinking statement & Descriptive sentence assigned from the principal event-time chain to a background layer. \\
插叙 & retrospective passage (analepsis) & Passage whose narrative time departs from the surrounding sequence and refers to an earlier point. \\
时间粒度 & temporal granularity & Precision level of a temporal expression, such as year, season or month. \\
跨文本事件核验与时间对齐 & cross-text event verification and temporal alignment & Verification that selected TimeBlocks refer to the same event or phase before source temporal evidence constrains the target TimeBlock. \\
证据链 & evidence chain & Traceable connection from a final chronological judgement to TimeBlocks, sentences and source texts. \\
\end{longtable}

\subsection{Texts and chapters}
Repeated text and chapter titles follow the forms in \suptabref{tab:supp-text-titles}.

\begin{longtable}{@{}P{0.23\textwidth}P{0.36\textwidth}P{0.33\textwidth}@{}}
\caption{Historical texts and chapter titles cited in the manuscript.}\label{tab:supp-text-titles}\\
\toprule
\textbf{Chinese title} & \textbf{English usage} & \textbf{Scope and note} \\
\midrule
\endfirsthead
\toprule
\textbf{Chinese title} & \textbf{English usage} & \textbf{Scope and note} \\
\midrule
\endhead
\bottomrule
\endfoot
史记 & \textit{Shiji} (\textit{Records of the Grand Historian}) & Historical text used in all method demonstrations; \textit{Shiji} is used after first definition. \\
\midrule
项羽本纪 & \textit{Basic Annals of Xiang Yu} & \textit{Shiji}, chapter 7; Xiang Yu's biographical perspective. \\
\midrule
高祖本纪 & \textit{Basic Annals of Gaozu} & \textit{Shiji}, chapter 8; Liu Bang's biographical perspective. \\
\midrule
萧相国世家 & \textit{Hereditary House of Chancellor Xiao} & \textit{Shiji}, chapter 53; Xiao He's biographical perspective. \\
\midrule
二十四史 & Twenty-Four Histories & Canonical collection of Chinese official histories in the annals-biographies format. \\
\midrule
三国史记 & \textit{Samguk Sagi} & Korean history written in the annals-biographies format. \\
\midrule
高丽史 & \textit{Goryeosa} & History of the Goryeo dynasty. \\
\midrule
大日本史 & \textit{Dai Nihonshi} & Japanese history composed within the Chinese historiographical tradition. \\
\midrule
梁启超年谱长编 & \textit{Extended Chronological Biography of Liang Qichao} & Major example of an extended biographical chronology for Liang Qichao. \\
\midrule
胡适年谱长编 & \textit{Extended Chronological Biography of Hu Shi} & Extended biographical chronology of Hu Shi. \\
\midrule
赫胥黎生平与书信集 & \textit{The Life and Letters of Thomas Henry Huxley} & Evidence-rich biographical and epistolary reference in the English-language tradition. \\
\midrule
希腊罗马名人传 & \textit{Parallel Lives} & Plutarch's comparative biographies. \\
\end{longtable}

\subsection{People, aliases and titles}
Historical names use pinyin in Chinese order, with surname first. Multiple source names and titles referring to the same person are grouped under one historical identity (\suptabref{tab:supp-person-names}).

\begin{longtable}{@{}P{0.17\textwidth}P{0.15\textwidth}P{0.23\textwidth}P{0.33\textwidth}@{}}
\caption{Historical people, aliases and titles used in the manuscript.}\label{tab:supp-person-names}\\
\toprule
\textbf{Person} & \textbf{Chinese form} & \textbf{English usage} & \textbf{Relation and note} \\
\midrule
\endfirsthead
\toprule
\textbf{Person} & \textbf{Chinese form} & \textbf{English usage} & \textbf{Relation and note} \\
\midrule
\endhead
\bottomrule
\endfoot
\multirow[t]{5}{=}{Liu Bang} & 刘邦 & Liu Bang & Personal name and default manuscript usage. \\
\cmidrule(l){2-4}
 & 高祖 & Gaozu & Temple name; \textit{Emperor Gaozu of Han} when imperial status is foregrounded. \\
\cmidrule(l){2-4}
 & 汉王 & King of Han & Royal title before the establishment of the Han empire. \\
\cmidrule(l){2-4}
 & 沛公 & Pei Gong & Early title in the uprising narrative; also translated as Duke or Lord of Pei. \\
\cmidrule(l){2-4}
 & 刘季 & Liu Ji & Early name. \\
\midrule
\multirow[t]{2}{=}{Xiang Yu} & 项羽 & Xiang Yu & Conventional name and default manuscript usage. \\
\cmidrule(l){2-4}
 & 项籍 & Xiang Ji & Personal name. \\
\midrule
Xiao He & 萧何 & Xiao He & Principal adviser and chancellor during Liu Bang's rise. \\
\midrule
Chen She & 陈涉 & Chen She & Leader of the Daze Village uprising, also known as Chen Sheng. \\
\midrule
Qin Shi Huang & 秦始皇 & Qin Shi Huang & First emperor of Qin; also \textit{the First Emperor of Qin}. \\
\midrule
Lv Gong & 吕公 & Lv Gong & Elder connected to Liu Bang through marriage; \textit{Lord Lv} translates the title. \\
\midrule
Liang Qichao & 梁启超 & Liang Qichao & Modern Chinese scholar whose discussion of the evidentiary role of biographical chronology is cited. \\
\midrule
Hu Shi & 胡适 & Hu Shi & Modern Chinese scholar whose extended chronology illustrates the compilation tradition. \\
\end{longtable}

\end{document}